\documentclass[11pt]{article}

\usepackage[margin=1.15in]{geometry}
\usepackage[utf8]{inputenc}
\usepackage[T1]{fontenc}
\usepackage{amsmath,amssymb,amstext}
\usepackage{bm}
\usepackage{newpxtext}
\usepackage[vvarbb]{newpxmath}
\usepackage{microtype}
\usepackage{graphicx}
\usepackage{booktabs}
\usepackage{enumitem}
\usepackage{xcolor}
\usepackage{array}
\usepackage{caption}
\usepackage{titlesec}
\usepackage{fancyhdr}
\usepackage[normalem]{ulem}
\usepackage{longtable}
\usepackage{fvextra}
\usepackage{calc}
\usepackage{newunicodechar}
\usepackage{soul}
\usepackage{hyperref}

\definecolor{headingcolor}{HTML}{1F3864}
\definecolor{linkcolor}{HTML}{2A5DB0}

\hypersetup{
  colorlinks=true,
  linkcolor=linkcolor,
  citecolor=linkcolor,
  urlcolor=linkcolor,
  pdftitle={Synchronous Verified Semantic Caching: An Empirical Study},
  pdfauthor={Chengyou Xin, Maryam Aly Mohamed Aly Abouelseoud}
}

\titleformat{\section}
  {\normalfont\sffamily\bfseries\large\color{headingcolor}}
  {\thesection}{0.6em}{}
\titleformat{\subsection}
  {\normalfont\sffamily\bfseries\color{headingcolor}}
  {\thesubsection}{0.6em}{}
\titlespacing*{\section}{0pt}{1.6em}{0.7em}
\titlespacing*{\subsection}{0pt}{1.2em}{0.4em}

\setlist[itemize]{itemsep=3pt,topsep=4pt}
\setlist[enumerate]{itemsep=3pt,topsep=4pt}

\fancypagestyle{plain}{
  \fancyhf{}
  \fancyfoot[C]{\small\sffamily\thepage}
  
}

\newunicodechar{—}{\textemdash}
\newunicodechar{–}{\textendash}
\newunicodechar{−}{\ensuremath{-}}
\newunicodechar{Δ}{\ensuremath{\Delta}}
\newunicodechar{→}{\ensuremath{\rightarrow}}
\newunicodechar{δ}{\ensuremath{\delta}}
\newunicodechar{τ}{\ensuremath{\tau}}
\newunicodechar{≈}{\ensuremath{\approx}}
\newunicodechar{≥}{\ensuremath{\geq}}
\newunicodechar{≤}{\ensuremath{\leq}}
\newunicodechar{×}{\ensuremath{\times}}
\newunicodechar{∈}{\ensuremath{\in}}
\newunicodechar{′}{\ensuremath{'}}
\newunicodechar{§}{\S}
\newunicodechar{…}{\ldots}

\providecommand{\tightlist}{\setlength{\itemsep}{0pt}\setlength{\parskip}{0pt}}
\makeatletter
\newsavebox\pandoc@box
\newcommand*\pandocbounded[1]{\sbox\pandoc@box{#1}%
  \ifdim\wd\pandoc@box>\linewidth\resizebox{\linewidth}{!}{\usebox\pandoc@box}\else\usebox\pandoc@box\fi}
\makeatother
\def\LTcaptype{none}
\usepackage{etoolbox}
\AtBeginEnvironment{longtable}{\small}

\title{\textbf{Synchronous Verified Semantic Caching:}\\[6pt]
  {\Large\normalfont An Empirical Study}}
\author{Chengyou Xin\\
  {\normalsize LoopDot AI Research}
  \and
  Maryam Aly Mohamed Aly Abouelseoud\\
  {\normalsize Alexandria University}}
\date{2026-07-26 (revised 2026-11-30)}

\begin{document}

\maketitle
\thispagestyle{plain}

\begin{abstract}
\noindent
Semantic caches reuse an LLM's past answer when a new query is similar enough to an old one, but similarity is not correctness: on the benchmarks studied here, a loose static threshold serves a wrong answer on 11--34\% of requests. Prior work either learns better thresholds (vCache) or adds an LLM judge that runs asynchronously, off the serving path, and is evaluated with an oracle in place of the judge (Krites). We measure the missing case: a real, lightweight verifier that checks gray-zone hits synchronously, before they are served, in a single-tier cache. We compare it with static and adaptive thresholds at matched error rate on two public benchmarks (about 210,000 requests), a third paraphrase dataset, and real customer-support traffic.

With an oracle verifier, synchronous gating raises hit rate by 20--28 percentage points at matched error rate, so the mechanism has real headroom. An off-the-shelf cross-encoder captures little of it: +0.9 to +3.1 points on LmArena under honest calibration, and a net loss on SearchQueries, where it ranks gray-zone candidates worse than raw similarity does. We call this a weak Go. Fine-tuning the same model on each dataset's own gray-zone labels raises held-out AUC from 0.72 to 0.88 and from 0.60 to 0.71, and the fine-tuned verifier beats the static threshold at all 18 honestly calibrated operating points across three datasets. It tolerates about 30\% label noise, but on multi-year support traffic a drifting base rate makes fine-tuning harmful; a simple change-point test on the labels detects this. Larger off-the-shelf rerankers, rewriting rejected answers, and verb-based bucketing do not substitute for fine-tuning. Finally, a blinded human check finds the benchmarks' correctness labels noisy at high similarity: comparisons at matched error rate are unaffected, but absolute error rates are overstated.
\end{abstract}

\begin{center}
\small\textbf{Code and full experimental artifacts:} \url{https://github.com/imxinchengyou/CacheVerifier}
\end{center}

{\small\noindent\textbf{Note:} An earlier technical report (Xin, 2026a) also covered formal risk control, adversarial robustness, and non-stationarity adaptation; those topics are treated in companion papers (Xin, 2026b, 2026c, 2026d). This paper is self-contained. Corrections made since the technical report's first release are summarized in Appendix A.}

\section{Introduction}\label{introduction}

Semantic caching, which replaces exact matching with vector similarity, is a standard technique for cutting the cost and latency of LLM applications. But similarity and answer correctness are different quantities, and a cache that serves a wrong answer is worse than one that misses.

Existing cache-hit logic is essentially \texttt{sim(q,\ h)\ ≥\ τ\ →\ hit}, where \texttt{sim} measures \emph{Question ≈ Question′}. The question that actually matters is whether \emph{Question → Answer} still holds, which is an answer-relevance problem, not a similarity problem. ``Can my dog eat honey'' and ``what's the deal with dogs and honey'' can score as highly similar and share an answer. ``How do I pause my subscription'' and ``how do I cancel my subscription'' also score as highly similar but need different answers, and a static threshold cannot tell the two cases apart.

Three published works, GPTCache, vCache, and Krites, each address part of this problem (Section 2). None has directly measured whether, in a single-tier cache with no offline curation and a real model as the verifier, it is worth putting verification on the serving path. That is this paper's question:

\begin{quote}
Under a single-tier cache architecture, can a real, lightweight verifier doing synchronous online gating achieve a higher hit rate than existing methods (static threshold, adaptive threshold) at a comparable error rate --- or a lower error rate at a comparable hit rate?
\end{quote}

This paper's contribution is purely empirical. Verifying cache hits is not itself new (Krites already uses an LLM judge, asynchronously and evaluated with an oracle); what is new is a systematic measurement of synchronous verification with real verifiers, which raises questions an asynchronous, oracle-based evaluation cannot answer: how often a real verifier is wrong, what it costs in latency, and what it takes to make it pay off. Concretely:

\begin{enumerate}
\tightlist
\item
  A reproducible evaluation pipeline (HNSW approximate nearest neighbor + a single-pass cached trace + cheap grid replay) that makes a full four-group sweep over roughly 210,000 real requests tractable. Under brute force, repeating the nearest-neighbor search once per grid point dominates the cost: about 19 hours for Groups A and B's grids alone. Under HNSW it collapses to a single pass of a few minutes. Verifier inference (Group D) and bootstrap confidence intervals remain separate costs that this optimization does not remove (Sections 5.4 and 5.5).
\item
  A line-by-line port of vCache's \texttt{VerifiedDecisionPolicy} (logistic-regression threshold estimation, delta-method variance, randomized exploration probability), not a simplified approximation.
\item
  Two measurements missing from the current literature: (a) the real (non-oracle) false-approve/false-reject rate of a verifier in a semantic-caching setting; (b) the measured latency that synchronous verification adds with a cross-encoder verifier.
\item
  An honest, interpolation-corrected Go/No-Go verdict --- the mechanism works but is highly verifier-dependent, rather than a blanket ``verification helps'' or ``verification doesn't help'' conclusion.
\end{enumerate}

\section{Related Work}\label{related-work}

{\def\LTcaptype{none} 
\begin{longtable}[]{@{}
  >{\raggedright\arraybackslash}p{(\linewidth - 8\tabcolsep) * \real{0.2000}}
  >{\raggedright\arraybackslash}p{(\linewidth - 8\tabcolsep) * \real{0.2000}}
  >{\raggedright\arraybackslash}p{(\linewidth - 8\tabcolsep) * \real{0.2000}}
  >{\raggedright\arraybackslash}p{(\linewidth - 8\tabcolsep) * \real{0.2000}}
  >{\raggedright\arraybackslash}p{(\linewidth - 8\tabcolsep) * \real{0.2000}}@{}}
\toprule\noalign{}
\begin{minipage}[b]{\linewidth}\raggedright
Work
\end{minipage} & \begin{minipage}[b]{\linewidth}\raggedright
Core method
\end{minipage} & \begin{minipage}[b]{\linewidth}\raggedright
Introduces a verifier?
\end{minipage} & \begin{minipage}[b]{\linewidth}\raggedright
Sync/async
\end{minipage} & \begin{minipage}[b]{\linewidth}\raggedright
Architectural precondition
\end{minipage} \\
\midrule\noalign{}
\endhead
\bottomrule\noalign{}
\endlastfoot
\textbf{GPTCache} (Bang, 2023) & Global fixed similarity threshold & No & --- & Single-tier \\
\textbf{vCache} (Schroeder et al., 2025, arXiv:2502.03771) & Online per-embedding threshold learning with an error-rate convergence guarantee & No & --- & Single-tier \\
\textbf{Krites} (2026, arXiv:2602.13165) & Static threshold unchanged; gray-zone matches trigger an \textbf{asynchronous} LLM judge that, on approval, promotes the answer into a dynamic tier & Yes, but only for the promotion decision & \textbf{Asynchronous, off the serving path} & Requires a two-tier architecture: an offline-curated static tier plus an online dynamic tier \\
\textbf{Closing the Operational Gap} (Baral et al., 2026, arXiv:2606.19719, EMNLP 2026 Industry Track; titled ``Closing the Calibration Gap'' in v1) & Calibrates the single similarity score already produced by the retrieval/reranking stage (P-CHR AUC / ORR metrics; ORR was called CRR in v1); adds no separate stage & No --- calibrates the one existing signal rather than adding a decoupled second-stage verifier & --- & Single-tier, one-signal architecture (better model selection and threshold calibration, but still ``score ≥ threshold → hit'') \\
\textbf{TweakLLM} (2025, arXiv:2507.23674) & On a match, skips the accept/reject decision entirely and uses a lightweight LLM to \textbf{dynamically rewrite} the cached answer to fit the new query & Partial --- substitutes generative rewriting for a binary verification decision & Synchronous (the rewrite happens on the serving path) & Single-tier, but reframes ``verification'' as ``editing'' \\
\textbf{Domain-specific embeddings} (Gill et al., 2025, arXiv:2504.02268) & Fine-tunes a compact embedding model for one epoch on domain data (plus a synthetic-data pipeline); reports higher precision and recall than open-source and proprietary embedding baselines & No --- improves the retrieval-stage similarity itself & --- & Single-tier, one-signal architecture \\
\textbf{Multi-turn semantic caching} (Atalar et al., 2026, arXiv:2611.01077) & Learns a per-pair ``reuse risk'' (probability a cached response is unacceptable) from LLM-judge labels on sampled reuses, uses it to choose a budgeted set of cached responses, and at request time serves a cached response only if its risk-weighted cost is below the generation cost; reports that embedding distance ranks acceptable reuse poorly in multi-turn conversations (AUC 0.62--0.74 on five workloads, vs.~0.75--0.89 for the learned risk) & Yes, a learned risk model (kernel ridge regression on pair features), not an LLM at serving time & Risk model used synchronously on the serving path; trained from logged judge labels & Multi-turn conversations (history-dependent reuse) \\
\end{longtable}
}

The last four rows are not direct competitors of Krites but adjacent lines of work: calibrating the existing signal, editing the cached content instead of verifying it, fine-tuning the retrieval embedding, and learning a reuse-risk model from judge feedback in multi-turn settings. Section 6.1 discusses Baral et al.~further. Gill et al.~reach, at the embedding stage, the same conclusion this paper reaches at the verification stage (Section 5.6): generic off-the-shelf models are not enough, and a short in-domain fine-tune is what helps; the two are complementary, since a better embedding changes which candidates reach the verifier, while the verifier decides whether a retrieved candidate may be served. Atalar et al.~likewise learn from feedback rather than relying on raw similarity, but target multi-turn conversations, which this paper does not evaluate (all of its datasets are single-turn; see Section 7). Two other papers from the same group study a different layer of the cache: which query-response pairs to keep under a size budget when query arrival probabilities and LLM serving costs are unknown and must be learned, with offline and online combinatorial-bandit algorithms and regret guarantees, first over a finite query set (Liu et al., 2026, arXiv:2508.07675) and then over a continuous embedding space (Atalar et al., 2026, arXiv:2604.20021). In both, the cost of reusing a cached response is modeled as a known non-decreasing function of embedding distance, so the serve-or-regenerate decision becomes a cost-derived similarity threshold; whether a specific retrieved answer is actually acceptable, the question this paper's verifier answers, is outside that model. The two layers are complementary: their algorithms decide what is in the cache, while the verifier decides whether a retrieved entry may be served.

Three points fix this paper's position relative to Krites specifically:

\begin{enumerate}
\tightlist
\item
  Krites explicitly discusses putting the judge directly on the serving path for blocking verification, and predicts that ``such a policy adds an extra model call to many requests, increasing both average and tail latency, eroding the cache's own benefit'' --- but \textbf{neither implements nor measures it}.
\item
  Krites' evaluation ``does not run the LLM judge in simulation\ldots{} but instantiates J directly from the benchmark's ground-truth equivalence-class relation,'' and its Discussion concedes that ``in production, an LLM-based verifier will have a non-zero false-reject and false-approve rate,'' but only offers an analytical upper bound --- \textbf{no empirical measurement}.
\item
  Krites' architectural precondition is a two-tier cache (an offline-curated static tier plus an online dynamic tier); its contribution is ``promoting a verified static answer into the dynamic tier,'' not ``judging in real time whether this hit can be trusted.'' The mechanism does not apply to a single-tier dynamic cache deployment with no offline curation pipeline.
\end{enumerate}

This paper's four experimental groups map directly onto those three gaps: Groups A and B reproduce existing baselines; Group C keeps Krites' own oracle verifier-fidelity assumption but makes verification synchronous, isolating the cost of ``going synchronous'' as a single variable; Group D swaps in a real verifier, isolating the cost of ``oracle → real'' as a single variable.

\section{Method}\label{method}

\subsection{Four experimental groups}\label{four-experimental-groups}

{\def\LTcaptype{none} 
\begin{longtable}[]{@{}
  >{\raggedright\arraybackslash}p{(\linewidth - 4\tabcolsep) * \real{0.3333}}
  >{\raggedright\arraybackslash}p{(\linewidth - 4\tabcolsep) * \real{0.3333}}
  >{\raggedright\arraybackslash}p{(\linewidth - 4\tabcolsep) * \real{0.3333}}@{}}
\toprule\noalign{}
\begin{minipage}[b]{\linewidth}\raggedright
Group
\end{minipage} & \begin{minipage}[b]{\linewidth}\raggedright
Method
\end{minipage} & \begin{minipage}[b]{\linewidth}\raggedright
Description
\end{minipage} \\
\midrule\noalign{}
\endhead
\bottomrule\noalign{}
\endlastfoot
A & Static threshold & \texttt{sim\ ≥\ τ\ →\ hit}, reproducing GPTCache; \texttt{τ} taken from the \texttt{STATIC\_\allowbreak{}THRESHOLDS} grid in vCache's own \texttt{benchmarks/\allowbreak{}benchmark.\allowbreak{}py} \\
B & Adaptive threshold & A line-by-line port of vCache's \texttt{VerifiedDecisionPolicy} (Section 3.2), not a simplified approximation \\
C & Synchronous + oracle verifier & The gray zone synchronously calls a ``perfect'' verifier (judging via ground-truth equivalence classes), but assigns it a non-zero modeled latency standing in for a real LLM judge (70ms default) \\
D & Synchronous + real verifier & The gray zone synchronously calls a real, non-oracle model \\
\end{longtable}
}

C and D share the same decision mechanism (\texttt{SynchronousVerifiedPolicy}). For a new query \(q\) whose nearest cached entry is \(h\), with cached answer \(a_h\) and similarity \(s = \mathrm{sim}(q, h)\), the cache serves \(a_h\) if and only if

\[
\mathrm{serve}(q) =
\begin{cases}
1, & s \ge \tau_{\mathrm{high}} \quad \text{(direct hit, no verifier call)} \\
\mathbb{1}\left[V(q, a_h) \ge \theta\right], & \tau_{\mathrm{low}} \le s < \tau_{\mathrm{high}} \quad \text{(gray zone, synchronous verifier call)} \\
0, & s < \tau_{\mathrm{low}} \quad \text{(miss, no verifier call)}
\end{cases}
\]

where \(V\) is the verifier's score and \(\theta\) its decision threshold. A request that is not served falls through to the LLM.

The only difference between C and D is the verifier implementation --- that is the point of the design: C isolates ``synchronous vs.~asynchronous'' (under Krites' own verifier-fidelity assumption), and D isolates ``oracle vs.~real model.''

\subsection{A faithful port of Group B}\label{a-faithful-port-of-group-b}

In vCache's official code (\texttt{vcache-project/\allowbreak{}vCache}, \texttt{vcache/\allowbreak{}vcache\_\allowbreak{}policy/\allowbreak{}strategies/\allowbreak{}verified.\allowbreak{}py}), \texttt{VerifiedDecisionPolicy}'s core logic is:

\begin{enumerate}
\tightlist
\item
  Each cache entry independently maintains its own observation history \texttt{(similarity,\ is\_\allowbreak{}correct)}, with \textbf{no global fallback across entries} --- with fewer than 6 observations (counting two synthetic seed observations every entry starts with; Section 5.2), the policy always declines to exploit (EXPLORE).
\item
  Once an entry reaches 6 observations, a 1-D logistic regression (similarity → correctness) is fit on its history, yielding a threshold estimate \texttt{t\_\allowbreak{}hat} and a slope \texttt{gamma}.
\item
  The variance \texttt{var\_\allowbreak{}t} of \texttt{t\_\allowbreak{}hat} is estimated via the delta method (or, under perfect separation, looked up in an empirical variance table shipped with vCache's own code).
\item
  Over a 50-point epsilon grid, a set of candidate confidence upper bounds \texttt{t\_\allowbreak{}prime} is computed and inverted into an ``explore probability'' \texttt{tau}, such that the expected error rate stays below the target \texttt{δ}.
\item
  A random draw \texttt{u\ \textasciitilde{}\ Uniform(0,1)} decides the action: explore (miss, verify, and update the observation history) if \texttt{u\ ≤\ tau}, otherwise exploit (hit, no verification).
\end{enumerate}

This paper's \texttt{AdaptiveThresholdPolicy} ports all five steps line by line, including the exact empirical variance-table values from vCache's own code, rather than substituting a simplified confidence-interval approximation.

\subsection{Computational efficiency: one pass, cheap replay}\label{computational-efficiency-one-pass-cheap-replay}

Because every request, hit or miss, is inserted into the cache (as in vCache's harness), the sequence of matched historical entries for Groups A/C/D is \textbf{entirely independent of the threshold parameters}. The evaluation pipeline exploits this by splitting the computation into three steps:

\begin{enumerate}
\tightlist
\item
  \texttt{build\_\allowbreak{}match\_\allowbreak{}trace}: one pass over the whole stream using HNSW (rather than brute force), recording each request's nearest-neighbor match --- computed exactly once.
\item
  \texttt{score\_\allowbreak{}gray\_\allowbreak{}zone}: the verifier scores, exactly once, every candidate whose similarity falls within the union of the entire parameter sweep.
\item
  \texttt{replay}: for each \texttt{(τ\_\allowbreak{}low,\ τ\_\allowbreak{}high,\ threshold)} combination, hit/miss decisions are cheaply re-derived from the cached results of the two steps above --- without touching the ANN index or the verifier model again.
\end{enumerate}

An empirical extrapolation of brute-force nearest-neighbor search (20,000 records, 1024 dimensions, 56 seconds per pass) shows that, without this optimization, repeating that search once per grid point across Groups A and B's threshold/delta grids on roughly 210,000 real records would take about 19 hours; with HNSW, the single shared pass collapses to minutes. This does not make the whole pipeline take minutes: Group D's verifier inference and the 200-resample bootstrap confidence intervals are separate costs unaffected by the ANN backend (Sections 5.4 and 5.5). It does remove what would otherwise be the dominant and prohibitive bottleneck. This fast path was cross-checked record-by-record against the slow path (\texttt{ExperimentRunner} + \texttt{SynchronousVerifiedPolicy}) run one grid point at a time, under identical parameters, and produced identical results.

\subsection{Ground truth for correctness}\label{ground-truth-for-correctness}

Following vCache's own harness: whether two records' answers are equivalent is judged by the dataset's own equivalence-class label (\texttt{ID\_\allowbreak{}Set} / \texttt{id\_\allowbreak{}set}), not by calling a real LLM to compare semantics. A hit's correctness is whether the matched historical entry shares the current request's equivalence class; on a miss, the same check (``would it have been correct had it hit'') is also performed, so every single request yields a full TP/FP/TN/FN entry in the confusion matrix.

\section{Experimental Setup}\label{experimental-setup}

\subsection{Datasets}\label{datasets}

{\def\LTcaptype{none} 
\begin{longtable}[]{@{}
  >{\raggedright\arraybackslash}p{(\linewidth - 6\tabcolsep) * \real{0.2500}}
  >{\raggedright\arraybackslash}p{(\linewidth - 6\tabcolsep) * \real{0.2500}}
  >{\raggedright\arraybackslash}p{(\linewidth - 6\tabcolsep) * \real{0.2500}}
  >{\raggedright\arraybackslash}p{(\linewidth - 6\tabcolsep) * \real{0.2500}}@{}}
\toprule\noalign{}
\begin{minipage}[b]{\linewidth}\raggedright
Dataset
\end{minipage} & \begin{minipage}[b]{\linewidth}\raggedright
Size
\end{minipage} & \begin{minipage}[b]{\linewidth}\raggedright
Source
\end{minipage} & \begin{minipage}[b]{\linewidth}\raggedright
Characteristics
\end{minipage} \\
\midrule\noalign{}
\endhead
\bottomrule\noalign{}
\endlastfoot
SemCacheLMArena & 60,000 records (drawn from a 63,796-record full set) & LM-Arena human-preference logs; GPT-4o-mini-generated, 3,500 classes with 1--23 paraphrases each & Conversational, long-form text, high lexical diversity \\
SemCacheSearchQueries & 150,000 records & ORCAS search logs, equivalence classes built via union-find + an LLM judge; the upstream \texttt{answer} field is an unpopulated constant placeholder, so answers were regenerated with the DeepSeek API for the 103,188 records the verifier reads & Short queries, keyword-dense \\
\end{longtable}
}

Both datasets come directly from HuggingFace \texttt{vCache/\allowbreak{}SemBenchmarkLmArena} and \texttt{vCache/\allowbreak{}SemBenchmarkSearchQueries}, using the datasets' own \textbf{precomputed embeddings} (\texttt{emb\_\allowbreak{}e5\_\allowbreak{}large\_\allowbreak{}v2} for LmArena, \texttt{emb\_\allowbreak{}gte} for SearchQueries) rather than re-encoding with sentence-transformers --- even with a matching model name, re-encoding risks failing to reproduce the paper's numbers due to version or pooling differences. All SearchQueries verifier results in this paper use the regenerated answers (Appendix A.1). Two further datasets are introduced where they are first used: Quora Question Pairs, a third, independently sourced paraphrase dataset (60,000 records; Section 5.6), and real customer-support traffic from two brands on Twitter (Section 5.8).

\subsection{Online evaluation protocol}\label{online-evaluation-protocol}

Following vCache's official harness: \textbf{no offline history split}. The cache starts empty, and the first \texttt{N} records of the dataset are streamed once, in their original order. Groups A/C/D require no pre-calibration; Group B's threshold learning also happens entirely online, within this same single pass. A 20\%/80\% history/evaluation split was considered and not used, so that results stay directly comparable to vCache's published numbers.

\subsection{Hyperparameter grids}\label{hyperparameter-grids}

\begin{itemize}
\tightlist
\item
  Group A's \texttt{τ}: \texttt{{[}0.\allowbreak{}80,\ 0.\allowbreak{}83,\ 0.\allowbreak{}86,\ 0.\allowbreak{}89,\ 0.\allowbreak{}92,\ 0.\allowbreak{}95,\ 0.\allowbreak{}97,\ 0.\allowbreak{}98,\ 0.\allowbreak{}99{]}} (matching vCache's \texttt{STATIC\_\allowbreak{}THRESHOLDS})
\item
  Group B's target error rate \texttt{δ}: \texttt{{[}0.\allowbreak{}01,\ 0.\allowbreak{}03,\ 0.\allowbreak{}05,\ 0.\allowbreak{}07{]}}, four points spanning vCache's \texttt{DELTAS} range (0.01--0.07); each Group B run refits per-entry thresholds often, so the full ten-point grid was not run
\item
  Group C/D's \texttt{τ\_\allowbreak{}high} fixed at \texttt{0.\allowbreak{}97} (one of Group A's own grid points, for comparability); \texttt{τ\_\allowbreak{}low} swept over \texttt{{[}0.\allowbreak{}80,\ 0.\allowbreak{}83,\ 0.\allowbreak{}86,\ 0.\allowbreak{}89,\ 0.\allowbreak{}92,\ 0.\allowbreak{}95{]}}
\item
  Group D's verifier raw-score decision threshold (\texttt{threshold}) is swept as a separate axis: \textbf{\texttt{{[}-2,\ -1,\ 0,\ 1,\ 2{]}} for LmArena, \texttt{{[}1,\ 3,\ 5,\ 7,\ 9,\ 11{]}} for SearchQueries} --- each chosen from that dataset's own score distribution, for the reason given in Section 5.4.
\end{itemize}

\subsection{Verifier implementations}\label{verifier-implementations}

\begin{itemize}
\tightlist
\item
  \textbf{Oracle} (Group C): approves based on ground-truth equivalence class, with a fixed 70ms modeled latency standing in for the API-class LLM judge (GPT-4.1-nano) Krites itself would call --- an explicitly labeled modeling assumption, not a measurement.
\item
  \textbf{Cross-Encoder} (Group D): \texttt{cross-encoder/\allowbreak{}ms-marco-MiniLM-L6-v2}, scoring the relevance of \texttt{(query,\ candidate\ cached\ answer)}, with latency measured as actual CPU inference time.
\end{itemize}

\subsection{Metrics}\label{metrics}

Over a stream of \(N\) requests, let \(\mathrm{hit}_i \in \{0,1\}\) indicate that request \(i\) was served from the cache and \(\mathrm{correct}_i \in \{0,1\}\) that the served or candidate answer is correct for it. The two headline metrics are

\[
H = \frac{1}{N}\sum_{i=1}^{N} \mathrm{hit}_i, \qquad
E = \frac{1}{N}\sum_{i=1}^{N} \mathrm{hit}_i\,(1-\mathrm{correct}_i),
\]

the hit rate and the error rate; the error rate's denominator is all requests, not only hits. Inside the gray zone, over the requests \(G\) on which the verifier is called, the false-approve rate is the share of approved candidates that are wrong, and the false-reject rate the share of rejected candidates that are right:

\[
\mathrm{FAR} = \frac{\#\{i \in G : \mathrm{hit}_i = 1,\ \mathrm{correct}_i = 0\}}{\#\{i \in G : \mathrm{hit}_i = 1\}}, \qquad
\mathrm{FRR} = \frac{\#\{i \in G : \mathrm{hit}_i = 0,\ \mathrm{correct}_i = 1\}}{\#\{i \in G : \mathrm{hit}_i = 0\}}.
\]

Also reported are precision and recall (treating ``was it served'' as a prediction of ``was the match correct''), the verifier call rate \(r = |G|/N\), the mean verifier latency \(\bar t_V\), and the expected added latency per request \(r \cdot \bar t_V\). Hit rate and error rate are reported with 95\% bootstrap confidence intervals (200 resamples).

\section{Results}\label{results}

\subsection{Group A: the classic static-threshold trade-off}\label{group-a-the-classic-static-threshold-trade-off}

\begin{figure}[htbp]
\centering
\includegraphics[width=0.92\linewidth,keepaspectratio]{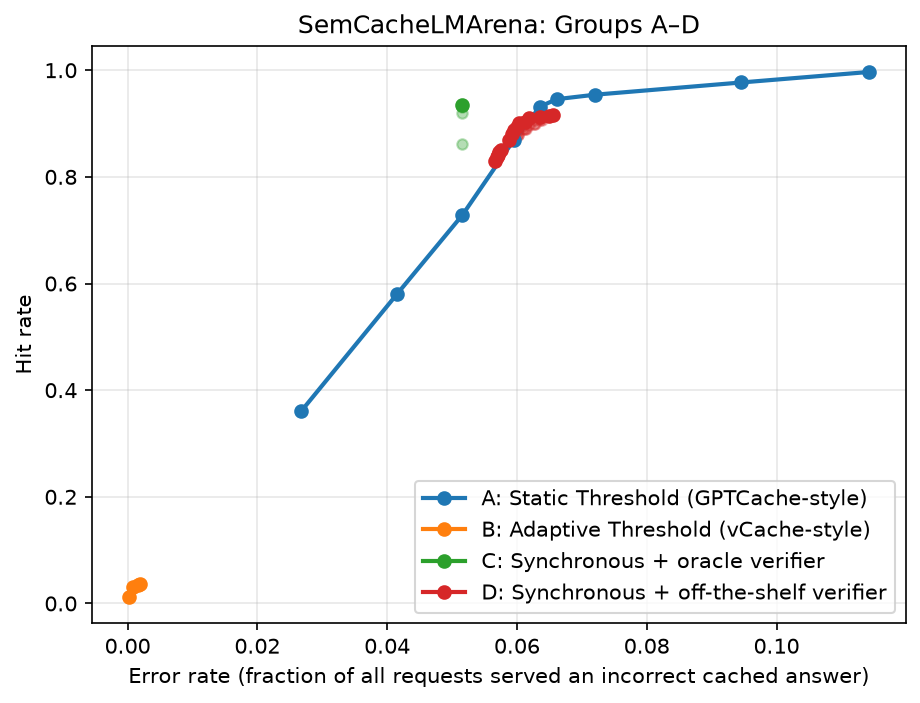}
\caption{Four-group Pareto comparison on SemCacheLMArena}
\end{figure}

\begin{figure}[htbp]
\centering
\includegraphics[width=0.92\linewidth,keepaspectratio]{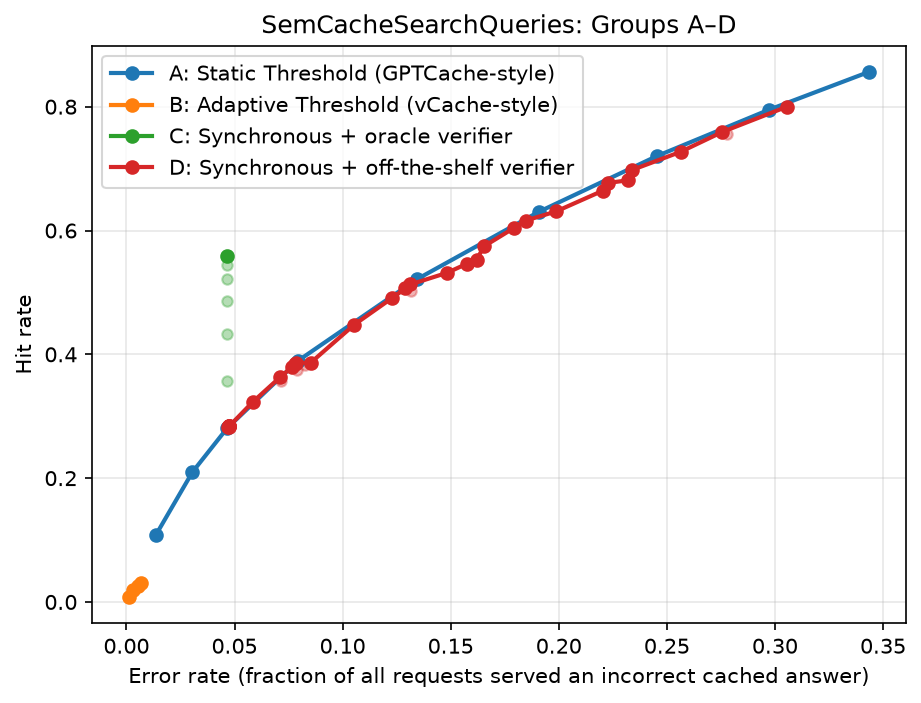}
\caption{Four-group Pareto comparison on SemCacheSearchQueries}
\end{figure}

Figures 1 and 2 plot Groups A--D together (Group B's points sit near the origin; Section 5.2); Group A's own curve is discussed first.

{\def\LTcaptype{none} 
\begin{longtable}[]{@{}
  >{\raggedright\arraybackslash}p{(\linewidth - 6\tabcolsep) * \real{0.2500}}
  >{\raggedright\arraybackslash}p{(\linewidth - 6\tabcolsep) * \real{0.2500}}
  >{\raggedright\arraybackslash}p{(\linewidth - 6\tabcolsep) * \real{0.2500}}
  >{\raggedright\arraybackslash}p{(\linewidth - 6\tabcolsep) * \real{0.2500}}@{}}
\toprule\noalign{}
\begin{minipage}[b]{\linewidth}\raggedright
Dataset
\end{minipage} & \begin{minipage}[b]{\linewidth}\raggedright
τ=0.80
\end{minipage} & \begin{minipage}[b]{\linewidth}\raggedright
τ=0.97
\end{minipage} & \begin{minipage}[b]{\linewidth}\raggedright
τ=0.99
\end{minipage} \\
\midrule\noalign{}
\endhead
\bottomrule\noalign{}
\endlastfoot
LmArena & hit 99.7\% / error \textbf{11.4\%} & hit 72.8\% / error 5.16\% & hit 36.1\% / error 2.68\% \\
SearchQueries & hit 85.6\% / error \textbf{34.3\%} & hit 28.1\% / error 4.66\% & hit 10.9\% / error 1.37\% \\
\end{longtable}
}

The error rate at a loose threshold (11.4\% on LmArena, 34.3\% on SearchQueries) confirms this paper's motivation (Section 1). SearchQueries' short, keyword-dense queries produce far more confusable pairs than LmArena's, so plain threshold tuning does worse on this kind of data.

\subsection{Group B: trading hit rate for a formal error-rate guarantee}\label{group-b-trading-hit-rate-for-a-formal-error-rate-guarantee}

{\def\LTcaptype{none} 
\begin{longtable}[]{@{}
  >{\raggedright\arraybackslash}p{(\linewidth - 8\tabcolsep) * \real{0.2000}}
  >{\raggedright\arraybackslash}p{(\linewidth - 8\tabcolsep) * \real{0.2000}}
  >{\raggedright\arraybackslash}p{(\linewidth - 8\tabcolsep) * \real{0.2000}}
  >{\raggedright\arraybackslash}p{(\linewidth - 8\tabcolsep) * \real{0.2000}}
  >{\raggedright\arraybackslash}p{(\linewidth - 8\tabcolsep) * \real{0.2000}}@{}}
\toprule\noalign{}
\begin{minipage}[b]{\linewidth}\raggedright
Dataset
\end{minipage} & \begin{minipage}[b]{\linewidth}\raggedright
δ=0.01
\end{minipage} & \begin{minipage}[b]{\linewidth}\raggedright
δ=0.03
\end{minipage} & \begin{minipage}[b]{\linewidth}\raggedright
δ=0.05
\end{minipage} & \begin{minipage}[b]{\linewidth}\raggedright
δ=0.07
\end{minipage} \\
\midrule\noalign{}
\endhead
\bottomrule\noalign{}
\endlastfoot
LmArena & 1.21\% / 0.02\% & 3.10\% / 0.08\% & 3.37\% / 0.15\% & 3.57\% / 0.20\% \\
SearchQueries & 0.80\% / 0.12\% & 1.96\% / 0.32\% & 2.64\% / 0.54\% & 3.06\% / 0.68\% \\
Quora & 0.12\% / 0.09\% & 0.39\% / 0.29\% & 0.58\% / 0.43\% & 0.91\% / 0.74\% \\
\end{longtable}
}

(Each cell gives hit rate / error rate over all requests; \texttt{results/\allowbreak{}*\_\allowbreak{}groupB\_\allowbreak{}seedfix.\allowbreak{}json}.)

The ported vCache \texttt{VerifiedDecisionPolicy} keeps the error rate far below the target δ at every point, and pays for it with a hit rate of at most a few percent. This is a structural cost of the algorithm: a cache entry can only be exploited after it has itself been the nearest neighbor of enough later requests, all served as misses, to fit its own threshold. vCache seeds every entry with two synthetic observations, \texttt{(similarity=0.\allowbreak{}0,\ incorrect)} and \texttt{(similarity=1.\allowbreak{}0,\ correct)}, so 4 real observations clear its \texttt{min\_\allowbreak{}observations=6} gate. When paraphrases within a semantic class spread their matches across different entries of that class, few entries accumulate even 4, and most stay in cold start. This is vCache's intended trade, a formal guarantee for hit rate, and not a porting error: the port's formulas (design matrix, gamma clipping, \texttt{t\_\allowbreak{}hat}, variance estimates, tau search, EXPLORE/EXPLOIT mapping, observing only on misses) and its seed observations were checked line by line against the official repository (\texttt{github.\allowbreak{}com/\allowbreak{}vcache-project/\allowbreak{}vCache}). Because Group B operates at a completely different hit-rate scale from A/C/D, it is not part of the Go/No-Go comparison below.

\subsection{Group C: the theoretical ceiling of the synchronous mechanism}\label{group-c-the-theoretical-ceiling-of-the-synchronous-mechanism}

Groups C and D are compared against a \textbf{linear interpolation} of Group A's Pareto frontier, rather than only against A's coarser discrete grid points, which would understate what A can achieve. Let \((E_k, H_k)\), \(k = 1, \dots, 9\), be Group A's grid points sorted by error rate, and \(\hat H_A(E)\) the piecewise-linear curve through them. The \textbf{net lead} of an operating point \((E, H)\) is

\[
\Delta = H - \hat H_A(E),
\]

the hit-rate gain over the static threshold at the same error rate. A point \emph{wins} (or \emph{loses}) when the bootstrap 95\% CI of \(\Delta\) lies entirely above (or below) zero, and \emph{ties} otherwise; ``best net lead'' is the largest \(\Delta\) over a grid. Points outside the range of Group A's error rates are not compared.

{\def\LTcaptype{none} 
\begin{longtable}[]{@{}
  >{\raggedright\arraybackslash}p{(\linewidth - 4\tabcolsep) * \real{0.3333}}
  >{\raggedright\arraybackslash}p{(\linewidth - 4\tabcolsep) * \real{0.3333}}
  >{\raggedright\arraybackslash}p{(\linewidth - 4\tabcolsep) * \real{0.3333}}@{}}
\toprule\noalign{}
\begin{minipage}[b]{\linewidth}\raggedright
Dataset
\end{minipage} & \begin{minipage}[b]{\linewidth}\raggedright
C's maximum lead over A's interpolated frontier
\end{minipage} & \begin{minipage}[b]{\linewidth}\raggedright
Do all grid points lead?
\end{minipage} \\
\midrule\noalign{}
\endhead
\bottomrule\noalign{}
\endlastfoot
LmArena & \textbf{+20.6 percentage points} (hit rate, at matched error rate) & Yes (6/6) \\
SearchQueries & \textbf{+27.9 percentage points} & Yes (6/6) \\
\end{longtable}
}

Under the assumption of a perfect verifier, the synchronous gating mechanism has substantial theoretical headroom on both datasets, and more so on SearchQueries --- because that dataset has more candidates that are ``high similarity but wrong,'' giving a verifier more room to recover. This shows the mechanism itself is not falsified: the question is not whether synchronous verification is worth doing in principle, but which verifier to use.

\subsection{Group D: a real verifier cashes in only a sliver of that ceiling}\label{group-d-a-real-verifier-cashes-in-only-a-sliver-of-that-ceiling}

The same interpolation-based comparison, applied to Group D (30 grid points for LmArena, 36 for SearchQueries):

{\def\LTcaptype{none} 
\begin{longtable}[]{@{}
  >{\raggedright\arraybackslash}p{(\linewidth - 8\tabcolsep) * \real{0.2000}}
  >{\raggedright\arraybackslash}p{(\linewidth - 8\tabcolsep) * \real{0.2000}}
  >{\raggedright\arraybackslash}p{(\linewidth - 8\tabcolsep) * \real{0.2000}}
  >{\raggedright\arraybackslash}p{(\linewidth - 8\tabcolsep) * \real{0.2000}}
  >{\raggedright\arraybackslash}p{(\linewidth - 8\tabcolsep) * \real{0.2000}}@{}}
\toprule\noalign{}
\begin{minipage}[b]{\linewidth}\raggedright
Dataset
\end{minipage} & \begin{minipage}[b]{\linewidth}\raggedright
Points beating A's interpolated frontier
\end{minipage} & \begin{minipage}[b]{\linewidth}\raggedright
Tied
\end{minipage} & \begin{minipage}[b]{\linewidth}\raggedright
Losing
\end{minipage} & \begin{minipage}[b]{\linewidth}\raggedright
Best net lead
\end{minipage} \\
\midrule\noalign{}
\endhead
\bottomrule\noalign{}
\endlastfoot
LmArena & 14/30 & 5/30 & 11/30 & \textbf{+1.9 percentage points} (hit rate; τ\_low=0.92, threshold=−1, error=5.99\%, hit=89.4\%) \\
SearchQueries & 1/36 & 12/36 & 23/36 & \textbf{+0.30 percentage points} (τ\_low=0.95, threshold=9, error=5.85\%, hit=32.33\%) \\
\end{longtable}
}

LmArena's best point is statistically significant: hit rate 89.4\% (95\% CI {[}89.2\%, 89.7\%{]}) versus Group A's nearest comparable point (τ=0.95, error rate 5.96\%): hit rate 87.0\% (95\% CI {[}86.7\%, 87.3\%{]}) --- the two confidence intervals do not overlap. But this lead is \textbf{less than one-tenth} of the oracle ceiling (+20.6pp), and more than a third of the 30 tested points (11) actually did \emph{worse} than simply tuning the static threshold, so on a fixed grid the benefit depends on the \texttt{(τ\_\allowbreak{}low,\ threshold)} chosen; the honest-calibration results below show what a calibrated threshold achieves.

The SearchQueries result is \textbf{net harmful}: only 1 of 36 grid points genuinely beats Group A's interpolated frontier, 12 tie, and \textbf{23 lose} to the static-threshold frontier; the best point's lead is only +0.30 percentage points. The gray zone's false-approve and false-reject rates range over 21\%--50\% and 33\%--74\% respectively. That is not the near-chance pattern of a model with no signal: the model is using the candidate answer's content, and is wrong often enough that the net effect is worse than having no verifier. This is also why this paper used different verifier-score threshold grids for the two datasets (Section 4.3): the grid calibrated on LmArena (\texttt{{[}-2,-1,0,1,2{]}}) applied to SearchQueries revealed that dataset's score distribution is \texttt{{[}-11.\allowbreak{}5,\ 11.\allowbreak{}6{]}}, mean ≈5.9 --- shaped quite differently from LmArena's, requiring a grid chosen from SearchQueries' own percentiles (\texttt{{[}1,3,5,7,9,11{]}}) rather than reused wholesale. Verifier hyperparameters do not transfer across domains.

\textbf{Honest-calibration ablation.} The verdict above sweeps a hand-picked, coarse threshold grid and reports the best point in it, which is optimistic because the threshold is chosen after seeing the results. An honest calibration procedure (Youden's J statistic, the same procedure the first author's commercial implementation uses) was run on all three datasets. At each tau\_low, the gray-zone stream is split chronologically into a calibration half and a test half. The threshold is picked by Youden's J on the calibration half only,

\[
\theta^{*} = \arg\max_{\theta} \left[\mathrm{TPR}(\theta) - \mathrm{FPR}(\theta)\right],
\]

where TPR and FPR are the verifier's true- and false-positive rates at threshold \(\theta\) on the calibration half, and hit rate and error rate are then measured on the test half. The result is compared against Group A's frontier on \textbf{exactly the same request set} (win/tie/loss by whether the bootstrap 95\% CI of the difference crosses 0; Appendix A.2 explains why the same request set matters):

{\def\LTcaptype{none} 
\begin{longtable}[]{@{}
  >{\raggedright\arraybackslash}p{(\linewidth - 2\tabcolsep) * \real{0.5000}}
  >{\raggedright\arraybackslash}p{(\linewidth - 2\tabcolsep) * \real{0.5000}}@{}}
\toprule\noalign{}
\begin{minipage}[b]{\linewidth}\raggedright
Dataset
\end{minipage} & \begin{minipage}[b]{\linewidth}\raggedright
Honest calibration vs.~Group A frontier on the same requests (6 tested points)
\end{minipage} \\
\midrule\noalign{}
\endhead
\bottomrule\noalign{}
\endlastfoot
SearchQueries (full 150,000 records) & 0 beat, 1 tie, \textbf{5 lost}, -0.01pp to -1.72pp \\
LmArena (full 60,000 records) & \textbf{6/6 beat}, +0.87pp to +3.10pp \\
Quora (full 60,000 records) & 0 beat, 3 tie, \textbf{3 lost}, -0.02pp to -0.46pp \\
\end{longtable}
}

\textbf{The ``net harmful'' verdict on SearchQueries holds under honest calibration.} Under an honestly-calibrated threshold, the same untuned, off-the-shelf verifier significantly loses to the static-threshold frontier on the same requests at 5 of 6 tested points and ties at 1, consistent with the 23 of 36 grid points lost above. More directly, inside the default gray zone, ranking by raw similarity beats this untuned cross-encoder at every error budget. Taking the top 5\% of gray-zone requests by each score, reuse is correct 0.732 of the time by similarity and 0.675 by the cross-encoder (difference 95\% CI {[}+0.038, +0.073{]}; \texttt{scripts/\allowbreak{}gray\_\allowbreak{}zone\_\allowbreak{}high\_\allowbreak{}precision\_\allowbreak{}compare.\allowbreak{}py}). On this dataset the off-the-shelf verifier judges the gray zone worse than simply raising the similarity threshold would. On LmArena, honest calibration beats the static threshold at all 6 tested points (+0.87pp to +3.10pp), consistent with the grid-search result (+1.9pp at the best point). Quora under honest calibration is likewise no better than the static threshold (3 lost, 3 tied). Diagnosis points to the dataset rather than the calibration method. The verifier's overall AUC is about 0.61, so there is real signal, but the score distributions of correct and incorrect candidates overlap heavily. QQP mixes annotation noise with deliberately constructed hard negatives that look alike on the surface but mean different things (e.g.~``George H.W. Bush'' vs.~``George W. Bush'' --- two different presidents, nearly identical text) that cap how separable any single threshold can make this dataset.

Taken together, the three datasets show that whether the untuned verifier beats the static threshold depends on whether its ranking inside that dataset's gray zone is better than similarity's own ranking --- yes on LmArena, no on SearchQueries and Quora; honest calibration removes the optimism of post-hoc grid search but does not change this picture.

\textbf{The same check for the fine-tuned verifier.} Section 5.6's Group E numbers (53/54 wins on SearchQueries) also come from post-hoc grid search. The identical procedure (chronological calibration/test split per tau\_low, Youden's J picked on the calibration half only) was rerun against each dataset's own Group E fine-tuned verifier (script \texttt{scripts/\allowbreak{}threshold\_\allowbreak{}calibration\_\allowbreak{}ablation\_\allowbreak{}groupE.\allowbreak{}py}; raw logs/JSON in \texttt{results/\allowbreak{}logs/\allowbreak{}} and \texttt{results/\allowbreak{}*\_\allowbreak{}groupE\_\allowbreak{}honest\_\allowbreak{}calibration.\allowbreak{}json}):

{\def\LTcaptype{none} 
\begin{longtable}[]{@{}
  >{\raggedright\arraybackslash}p{(\linewidth - 6\tabcolsep) * \real{0.2500}}
  >{\raggedright\arraybackslash}p{(\linewidth - 6\tabcolsep) * \real{0.2500}}
  >{\raggedright\arraybackslash}p{(\linewidth - 6\tabcolsep) * \real{0.2500}}
  >{\raggedright\arraybackslash}p{(\linewidth - 6\tabcolsep) * \real{0.2500}}@{}}
\toprule\noalign{}
\begin{minipage}[b]{\linewidth}\raggedright
Dataset
\end{minipage} & \begin{minipage}[b]{\linewidth}\raggedright
Honest calibration vs.~Group A frontier on the same requests (6 tested points)
\end{minipage} & \begin{minipage}[b]{\linewidth}\raggedright
Honest calibration best net lead
\end{minipage} & \begin{minipage}[b]{\linewidth}\raggedright
Post-hoc grid search (Section 5.6)
\end{minipage} \\
\midrule\noalign{}
\endhead
\bottomrule\noalign{}
\endlastfoot
LmArena & \textbf{6/6 beat} & +6.92pp & compared against Group D, not Group A (Section 5.6) \\
SearchQueries & \textbf{6/6 beat} & +2.33pp & 53/54 beat, 1 tie, 0 lost, +4.50pp \\
Quora & \textbf{6/6 beat} & +0.87pp & 19/48 beat, 29/48 tie, 0/48 lost, +2.03pp \\
\end{longtable}
}

The verdict: \textbf{Group E's core claim, that the fine-tuned verifier is never a net loss against the static-threshold frontier, holds fully under honest calibration}, and is in fact stronger: all 18 tested points across the three datasets significantly beat the static threshold. In magnitude, LmArena's best honest-calibration lead (+6.92pp) is similar to or above the grid-search numbers; SearchQueries' is clearly below the grid-search figure (+2.33pp vs.~+4.50pp); Quora narrows (+0.87pp vs.~+2.03pp, traceable to the QQP score-separability ceiling diagnosed earlier in this section rather than a calibration-method artifact). One caveat: honest calibration yields one test point per tau\_low, while grid search sweeps a whole threshold grid per tau\_low, so ``6/6'' and ``53/54'' have different denominators. What is comparable is the direction (win/tie/loss proportions) and the size of the lead when winning, not the ratios themselves. A second caveat concerns provenance: the LmArena and SearchQueries checkpoints are the same ones behind Section 5.6 (LmArena's is on Hugging Face as \texttt{ChengyouXin/\allowbreak{}cacheverifier-lmarena}). Quora's original checkpoint was not kept, so it was retrained from the same train/test split (\texttt{results/\allowbreak{}finetune\_\allowbreak{}verifier\_\allowbreak{}experiment\_\allowbreak{}quora.\allowbreak{}examples.\allowbreak{}json}, 11,928 train / 5,112 test). The untuned AUC reproduces exactly (0.6309) and the retrained fine-tuned AUC is within 0.0014 of the original (0.7407 vs.~0.7393): a faithful replication, though not literally the same weights.

\subsection{The latency cost of going synchronous}\label{the-latency-cost-of-going-synchronous}

Synchronous verification adds the verifier's latency to every gray-zone request, so the expected added latency per request is \(r \cdot \bar t_V\), the gray-zone call rate times the verifier's mean latency (Section 4.5). Krites' paper notes that a synchronous judge ``will increase latency'' without measuring it; this section measures it for the cross-encoder verifier.

The hosted \texttt{/\allowbreak{}v1/\allowbreak{}verify} service runs \texttt{ms-marco-MiniLM-L6-v2} on a small CPU-only server. It was measured three ways: the server-reported forward pass over 1,700 real gray-zone requests from AmazonHelp (Section 5.8); a 300-call loopback probe from the server to itself; and 150 calls from a client in the eastern US.

{\def\LTcaptype{none} 
\begin{longtable}[]{@{}
  >{\raggedright\arraybackslash}p{(\linewidth - 6\tabcolsep) * \real{0.2500}}
  >{\raggedright\arraybackslash}p{(\linewidth - 6\tabcolsep) * \real{0.2500}}
  >{\raggedright\arraybackslash}p{(\linewidth - 6\tabcolsep) * \real{0.2500}}
  >{\raggedright\arraybackslash}p{(\linewidth - 6\tabcolsep) * \real{0.2500}}@{}}
\toprule\noalign{}
\begin{minipage}[b]{\linewidth}\raggedright
Measurement
\end{minipage} & \begin{minipage}[b]{\linewidth}\raggedright
Mean
\end{minipage} & \begin{minipage}[b]{\linewidth}\raggedright
Median
\end{minipage} & \begin{minipage}[b]{\linewidth}\raggedright
p95
\end{minipage} \\
\midrule\noalign{}
\endhead
\bottomrule\noalign{}
\endlastfoot
Forward pass only (n=1,700) & --- & 17.5ms & 32.0ms \\
Full server stack, no network (loopback) & --- & \textasciitilde33ms & \textasciitilde44ms \\
End to end from a US-East client (warm connection) & 33.5ms & 32.8ms & 39.1ms \\
\end{longtable}
}

The end-to-end figure nearly matches the no-network loopback figure, so on that path the network adds little. (The cross-encoder latencies logged during the Section 5.4 experiments, about 66--68ms on SearchQueries and 288ms on LmArena, were measured while the machine ran other jobs and are not used.) Expected added latency at the highest-call-rate points (τ\_low=0.80), with the oracle's modeled 70ms (Section 4.4) shown for comparison:

{\def\LTcaptype{none} 
\begin{longtable}[]{@{}
  >{\raggedright\arraybackslash}p{(\linewidth - 6\tabcolsep) * \real{0.2500}}
  >{\raggedright\arraybackslash}p{(\linewidth - 6\tabcolsep) * \real{0.2500}}
  >{\raggedright\arraybackslash}p{(\linewidth - 6\tabcolsep) * \real{0.2500}}
  >{\raggedright\arraybackslash}p{(\linewidth - 6\tabcolsep) * \real{0.2500}}@{}}
\toprule\noalign{}
\begin{minipage}[b]{\linewidth}\raggedright
Dataset
\end{minipage} & \begin{minipage}[b]{\linewidth}\raggedright
Gray-zone call rate
\end{minipage} & \begin{minipage}[b]{\linewidth}\raggedright
Cross-encoder (\textasciitilde33ms)
\end{minipage} & \begin{minipage}[b]{\linewidth}\raggedright
Oracle, 70ms modeled
\end{minipage} \\
\midrule\noalign{}
\endhead
\bottomrule\noalign{}
\endlastfoot
SearchQueries & 57.5\% & \textasciitilde19ms & 40.3ms \\
Quora & 28.4\% & \textasciitilde9ms & 19.9ms \\
LmArena & 26.8\% & \textasciitilde9ms & 18.8ms \\
\end{longtable}
}

These recompute only latency; no hit-rate or error-rate number changes. A co-located cross-encoder therefore adds about 9--19ms per request. The oracle's 70ms is a modeling assumption for an API-hosted LLM judge, not a measurement; a real LLM judge's latency depends on the model, the provider and the network path, and was not measured here. The cross-encoder was measured from only one client region, in a single run, without concurrent load (Section 7).

\subsection{Group E: a domain-fine-tuned verifier narrows the gap on all three datasets}\label{group-e-a-domain-fine-tuned-verifier-narrows-the-gap-on-all-three-datasets}

Section 5.4's diagnosis was that the off-the-shelf cross-encoder's benefit is fragile on LmArena and negative on SearchQueries, because the model was never calibrated to either dataset's own gray zone. The direct test is to fine-tune the same base model (\texttt{cross-encoder/\allowbreak{}ms-marco-MiniLM-L6-v2}) separately on each dataset's own gray-zone examples, which are \texttt{(query,\ candidate\ cached\ answer)} pairs labelled by whether the hit would have been correct, and measure how much of Group D's shortfall against Group C closes.

\textbf{Setup.} Each dataset's gray zone (\texttt{τ\_\allowbreak{}low=0.\allowbreak{}80,\ τ\_\allowbreak{}high=0.\allowbreak{}97}) is split \textbf{by stream position, not randomly}: the first 70\% for fine-tuning and the last 30\% held out for evaluation. This mirrors calibrating on past traffic and deploying on future traffic, rather than i.i.d. cross-validation. LmArena yields 16,102 labeled examples (11,271 train / 4,831 test); SearchQueries, being shorter-text and higher-call-rate (Section 5.5), yields far more: 86,284 (60,398 train / 25,886 test). The base model is fine-tuned for one epoch with \texttt{sentence-transformers}' \texttt{CrossEncoderTrainer} and a binary cross-entropy loss, on CPU --- LmArena's longer conversational text took 43 minutes to fine-tune; SearchQueries' short keyword queries, despite 5.4× more training examples, took under 6 minutes, since per-example compute scales with sequence length.

\textbf{Held-out AUC.}

{\def\LTcaptype{none} 
\begin{longtable}[]{@{}llll@{}}
\toprule\noalign{}
Dataset & Untuned AUC & Fine-tuned AUC & Δ \\
\midrule\noalign{}
\endhead
\bottomrule\noalign{}
\endlastfoot
LmArena & 0.7212 & 0.8789 & \textbf{+0.158} \\
SearchQueries & 0.5984 & 0.7120 & \textbf{+0.114} \\
\end{longtable}
}

LmArena's untuned verifier already had some signal (0.72); fine-tuning sharpens it. SearchQueries' untuned verifier has AUC 0.5984 --- also real signal, but as Section 5.4 shows, that signal is frequently used to make the wrong call, so the net effect is worse than no verifier at all. Fine-tuning does not fully repair the absolute gap to LmArena (0.7120 vs.~0.88), but it moves the model from \emph{has signal, frequently misuses it} to \emph{has real signal and mostly uses it correctly}.

\textbf{Effect on the Pareto frontier.} Plugging each fine-tuned model back into the same synchronous gray-zone gating mechanism (\texttt{SynchronousVerifiedPolicy}, unchanged) and replaying it over the full stream as Group E, then comparing against Group A's interpolated frontier --- the same test Section 5.4 applied to Group D:

{\def\LTcaptype{none} 
\begin{longtable}[]{@{}
  >{\raggedright\arraybackslash}p{(\linewidth - 4\tabcolsep) * \real{0.3333}}
  >{\raggedright\arraybackslash}p{(\linewidth - 4\tabcolsep) * \real{0.3333}}
  >{\raggedright\arraybackslash}p{(\linewidth - 4\tabcolsep) * \real{0.3333}}@{}}
\toprule\noalign{}
\begin{minipage}[b]{\linewidth}\raggedright
Dataset
\end{minipage} & \begin{minipage}[b]{\linewidth}\raggedright
Group D vs.~A (Section 5.4)
\end{minipage} & \begin{minipage}[b]{\linewidth}\raggedright
Group E vs.~A
\end{minipage} \\
\midrule\noalign{}
\endhead
\bottomrule\noalign{}
\endlastfoot
LmArena & 14/30 beat, 11/30 lost, best +1.9pp & hit rate \textbf{+2.8 to +8.3pp} vs.~Group D itself at matched error rate; frontier strictly wider than D's \\
SearchQueries & \textbf{1/36 beat, 12/36 tied, 23/36 lost}, best +0.30pp & \textbf{53/54 beat, 1/54 tied, 0/54 lost}, best net lead \textbf{+4.50pp} (τ\_low=0.83, threshold=0.46, error=10.80\%, hit=50.32\%) \\
\end{longtable}
}

\begin{figure}[htbp]
\centering
\includegraphics[width=0.92\linewidth,keepaspectratio]{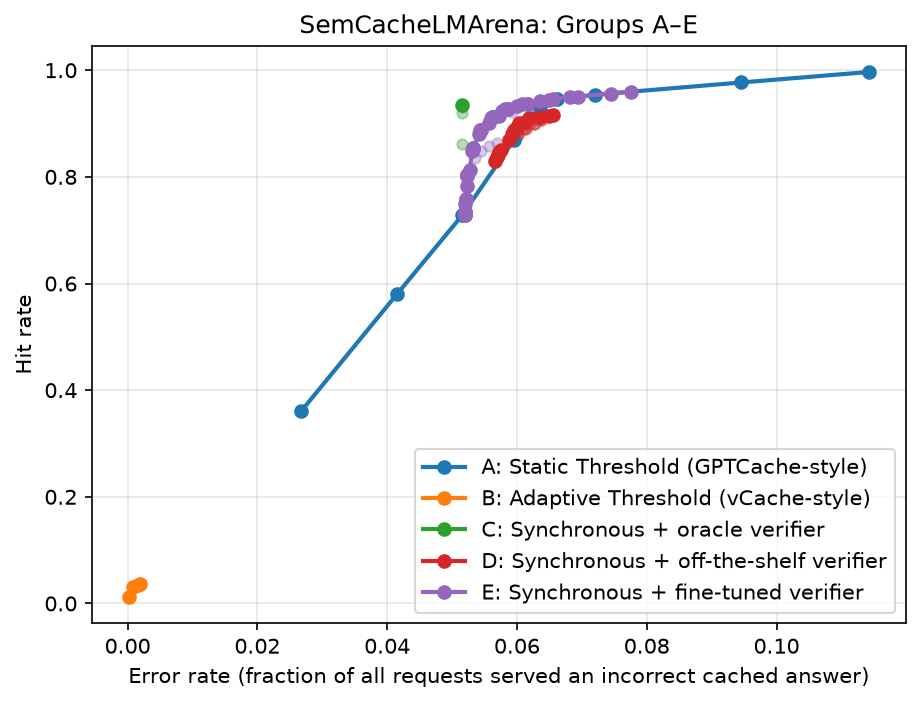}
\caption{Group E (fine-tuned verifier) added to the four-group Pareto comparison on SemCacheLMArena}
\end{figure}

\begin{figure}[htbp]
\centering
\includegraphics[width=0.92\linewidth,keepaspectratio]{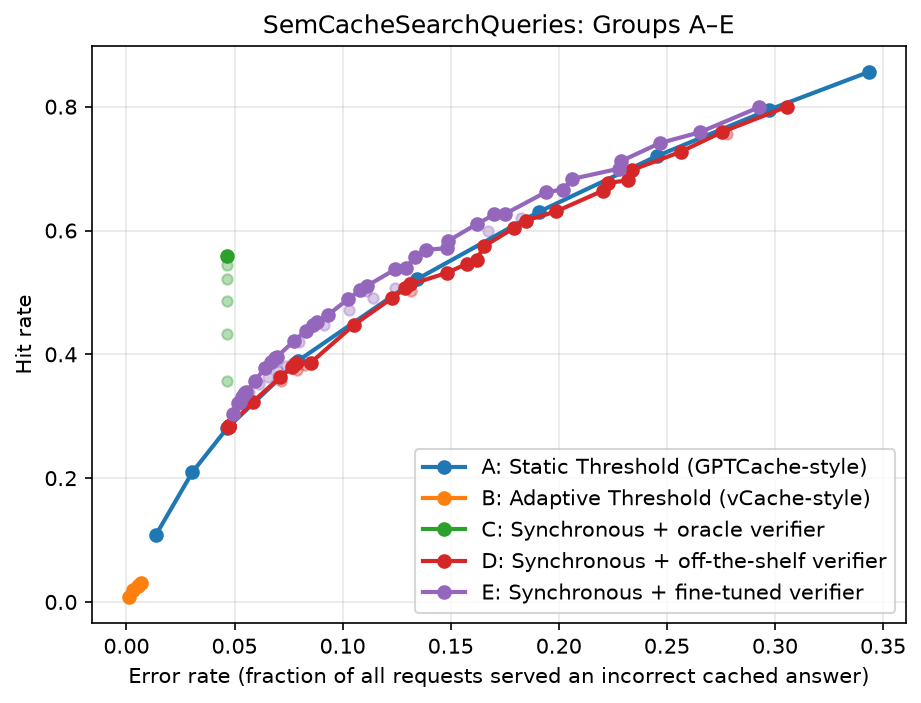}
\caption{Group E (fine-tuned verifier) added to the four-group Pareto comparison on SemCacheSearchQueries}
\end{figure}

\begin{figure}[htbp]
\centering
\includegraphics[width=0.92\linewidth,keepaspectratio]{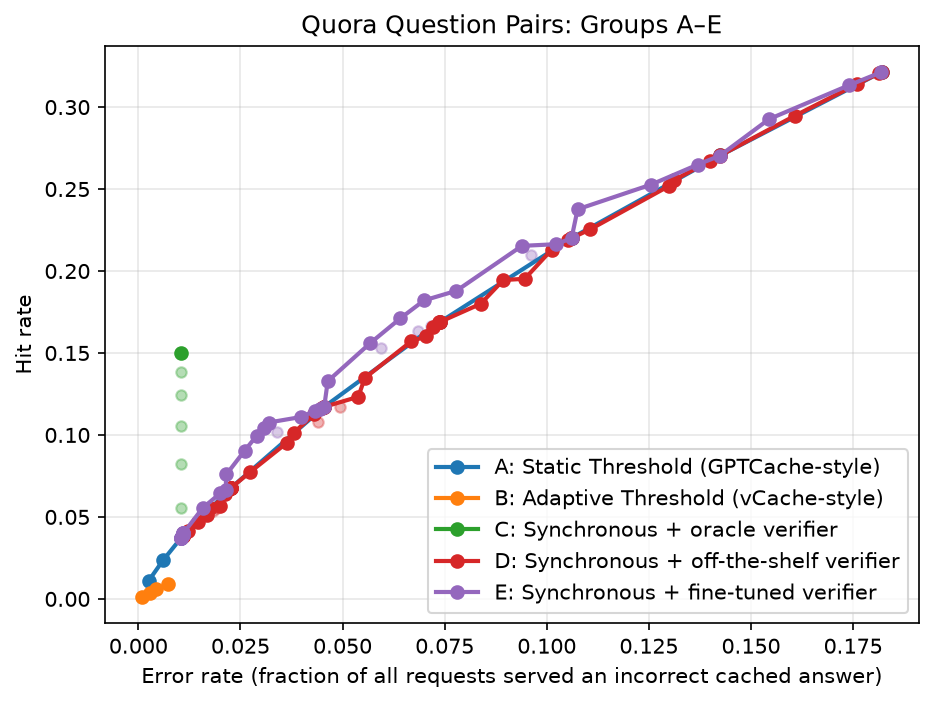}
\caption{Group E (fine-tuned verifier) added to the four-group Pareto comparison on Quora Question Pairs --- a third, independently-sourced dataset}
\end{figure}

The two datasets tell different but complementary stories. On LmArena, fine-tuning turns a fragile, partly-losing verifier into one that strictly dominates the static-threshold frontier and widens the achievable region in both directions (error rate down to \textasciitilde0.052, hit rate up to \textasciitilde0.96). On SearchQueries, fine-tuning turns a verifier that was \textbf{net harmful} (23 of 36 operating points lost to the static threshold) into one that beats the static baseline at 53 of 54 tested operating points, ties at 1, and loses at zero (best net lead +4.50pp), even though its absolute AUC (0.7120) remains well below LmArena's (0.88).

This is the clearest evidence in this paper that Section 6.1's diagnosis (``the variable that matters is how well the verifier matches the data domain'') is actionable, not just descriptive: the same base architecture, given in-domain labeled gray-zone examples, converts a fragile or net-harmful Group D result into a robust improvement over the static-threshold frontier on all three datasets tested. Here the labels come from the benchmarks' own equivalence classes; a production deployment would have to collect them through audits or user feedback, which are noisier (Section 5.7 tests how much noise the recipe tolerates).

\textbf{A third, independently-sourced check.} To rule out that this pattern is specific to vCache's own two benchmarks, the same recipe was repeated on Quora Question Pairs (GLUE's mirror) --- 60,000 short, real user-submitted questions with human-annotated duplicate labels, a domain neither vCache nor this paper's earlier experiments touched. Quora has no LLM-generated answers or native equivalence-class column. Unlike LmArena and SearchQueries, its equivalence classes were reconstructed by union-find over duplicate-labelled pairs, and \texttt{answer\ =\ query} (a past matched question's own text stands in for the cached response); details are in \texttt{scripts/\allowbreak{}convert\_\allowbreak{}quora\_\allowbreak{}dataset.\allowbreak{}py}. Quora's oracle ceiling is markedly lower than the other two datasets (+11.3pp over Group A's frontier, vs.~+20.6pp on LmArena and +27.9pp on SearchQueries), because only about a third of Quora's questions belong to a multi-member duplicate class --- most questions here are simply unique, capping how much any verifier could ever help.

Group D's off-the-shelf verifier again adds essentially nothing (0/54 beat, 50/54 tied, 4/54 \emph{lost} to the static-threshold frontier, best net lead +0.11pp). Its held-out AUC (0.6309) is above chance and close to SearchQueries' 0.5984: some discriminative power, but not enough to reliably beat an already-strong similarity baseline. Fine-tuning (11,928 in-domain examples) raises AUC to 0.7393 (+0.108) and, plugged back in as Group E, turns that null result into a modest but unambiguous win: 19/48 grid points beat Group A's frontier, 29/48 tied, \textbf{0/48 lost} (best net lead +2.03pp; +0.76pp mean improvement over Group D at matched error rate). The margin is the smallest of the three datasets, in line with Quora's smaller headroom. The qualitative pattern holds on all three datasets: fine-tuning \textbf{never produced a worse result than the untuned verifier}, and turned every fragile, null, or net-harmful Group D result into at least a modest, loss-free improvement over the static threshold.

\subsection{Practical deployment considerations: label noise, cold start, and drift}\label{practical-deployment-considerations-label-noise-cold-start-and-drift}

Section 5.6 fine-tunes on clean, oracle-quality labels harvested from Groups C/D's own ground truth. Three further questions decide whether this works in production rather than only in the lab. (1) How much label noise, from a realistic feedback signal such as user thumbs-up/down rather than an oracle, can fine-tuning tolerate before it stops helping or starts hurting? (2) How many labelled examples does a new deployment need before turning fine-tuning on is worth the risk? (3) How quickly does a fine-tuned verifier's advantage decay as traffic drifts away from its training window, which sets how often it must be refreshed? Each was tested on all three datasets, reusing the same fine-tuning recipe as Section 5.6, on an NVIDIA T4 GPU.

\textbf{Label noise robustness.} Training labels in a 4,000-example fixed subset were randomly flipped at six noise levels (0/5/10/20/30/40\%) before fine-tuning (one epoch); test labels were never touched, so held-out AUC always measures against true ground truth.

{\def\LTcaptype{none} 
\begin{longtable}[]{@{}llll@{}}
\toprule\noalign{}
Noise & LmArena Δ & SearchQueries Δ & Quora Δ \\
\midrule\noalign{}
\endhead
\bottomrule\noalign{}
\endlastfoot
0\% & +0.107 & +0.017 & +0.083 \\
5\% & +0.095 & +0.015 & +0.078 \\
10\% & +0.083 & +0.016 & +0.074 \\
20\% & +0.051 & +0.015 & +0.044 \\
30\% & +0.028 & +0.003 & +0.006 \\
40\% & \textbf{-0.101} & \textbf{-0.011} & \textbf{-0.037} \\
\end{longtable}
}

All three datasets show the same shape: a gentle, roughly linear decline through 30\% noise, then a rapid approach to, or across, zero. The interpolated zero-crossing is \textbf{32.2\% for LmArena, ≈32.1\% for SearchQueries, and 31.4\% for Quora}. Below roughly 30\% label noise, fine-tuning reliably helps; above it, all three datasets learn the wrong association and end up worse than not fine-tuning at all.

\textbf{Cold-start data requirements.} The same recipe was re-run with increasingly larger prefixes of each dataset's own (temporally ordered) training split, from 50 examples up to the full available set (three epochs, to give small subsets a fair chance to converge).

{\def\LTcaptype{none} 
\begin{longtable}[]{@{}llll@{}}
\toprule\noalign{}
Train size & LmArena Δ & SearchQueries Δ & Quora Δ \\
\midrule\noalign{}
\endhead
\bottomrule\noalign{}
\endlastfoot
50 & +0.001 & \textbf{-0.013} & +0.012 \\
200 & +0.030 & \textbf{-0.017} & +0.018 \\
500 & +0.050 & \textbf{-0.008} & +0.002 \\
1,000 & +0.080 & +0.004 & +0.049 \\
4,000 & +0.136 & +0.029 & +0.096 \\
Full (11.3k--60.4k) & +0.188 & +0.143 & +0.118 \\
\end{longtable}
}

All three curves rise smoothly on a log scale with \textbf{no saturation} even at full dataset size: more in-domain data keeps helping. The small-sample region is not uniformly benign, though. On SearchQueries, fine-tuning hurts through 500 examples and only turns positive at 1,000; LmArena's stronger baseline shows no such zone at the same sizes. A plausible reading is that SearchQueries' untuned verifier already carries some signal (AUC 0.5984), which small-sample fine-tuning can scramble before it has enough data to improve on it.

\textbf{Temporal drift and re-tuning cadence.} Each dataset's full gray-zone stream (train+test recombined, resorted by original position) was split into 8 equal-sized sequential chunks. A model fine-tuned on chunk \(i\) (the ``anchor'') was frozen and evaluated on every chunk \(j \geq i\); delta (vs.~that chunk's own untuned baseline) was averaged across all anchors at each temporal distance \(j-i\).

{\def\LTcaptype{none} 
\begin{longtable}[]{@{}llll@{}}
\toprule\noalign{}
Distance & LmArena Δ & SearchQueries Δ & Quora Δ \\
\midrule\noalign{}
\endhead
\bottomrule\noalign{}
\endlastfoot
0 (same chunk) & +0.193 & +0.308 & +0.298 \\
1 & +0.111 & +0.082 & +0.089 \\
4 & +0.124 & +0.075 & +0.085 \\
7 (farthest) & +0.129 & +0.056 & +0.069 \\
\end{longtable}
}

On all three datasets, the decay happens almost entirely in the first step. The gain drops sharply from distance 0 to distance 1 (the same-chunk figure partly reflects memorization), and that first step accounts for about 90\% or more of the whole decline from distance 0 to distance 7. After it, the gain levels off and stays clearly positive out to the farthest distance: +0.111 to +0.129 on LmArena, +0.089 to +0.069 on Quora, and +0.082 to +0.056 on SearchQueries. None of the three shows anything like the continued decay comcastcares shows in Section 5.8.

\textbf{Implication.} Label noise up to about 30\% is tolerable; the minimum amount of in-domain data depends on the dataset (about 1,000 examples on SearchQueries, fewer on LmArena and Quora); and a one-time fine-tune loses most of its same-window advantage in the first step and then levels off. The risk that needs engineering is feedback-signal quality: a gate that estimates a deployment's actual label-noise rate (e.g.~against a small manually-audited sample) before enabling fine-tuning is not optional, and the minimum-data floor should be set more conservatively where the untuned verifier is weak.

\subsection{A further check: real production customer-support traffic, and a genuine counter-example}\label{a-further-check-real-production-customer-support-traffic-and-a-genuine-counter-example}

Sections 5.6--5.7 validate the fine-tuning recipe on three datasets, but all three are public research benchmarks, not real business traffic. As a further check, the same recipe was applied to two brands' streams from the Kaggle ``Customer Support on Twitter'' dataset (Axelbrooke, 2017; CC0) --- 3M+ real support tweets from dozens of companies' official accounts, with genuine per-tweet timestamps and full thread structure (\texttt{tweet\_\allowbreak{}id}, \texttt{in\_\allowbreak{}response\_\allowbreak{}to\_\allowbreak{}tweet\_\allowbreak{}id}, \texttt{response\_\allowbreak{}tweet\_\allowbreak{}id}, \texttt{created\_\allowbreak{}at}). Two brands were selected to match e-commerce and telecom use cases specifically: \textbf{AmazonHelp} (60,000 query-answer pairs, 2015-06 to 2017-10) and \textbf{comcastcares} (30,180 pairs, 2014-07 to 2017-12 --- a 3.5-year span, the longest of any dataset in this paper).

\textbf{Construction.} Neither brand's stream has an LLM-generated answer or a native equivalence-class label. Each company reply tweet is paired with the customer tweet it answers (\texttt{query}, \texttt{answer}). There is no ground truth for whether two customer questions are the same issue, so equivalence classes are reconstructed from a behavioral signal: whether the company sent near-duplicate replies. Replies are clustered by sentence-embedding cosine similarity (\texttt{sentence-transformers/\allowbreak{}all-MiniLM-L6-v2}, threshold 0.92, \texttt{sentence\_\allowbreak{}transformers.\allowbreak{}util.\allowbreak{}community\_\allowbreak{}detection}) rather than exact string equality, because canned replies are often personalized with a customer name or order number. Exact string matching was not used: it under-counted true duplicates so severely that Group A's false-accept rate approached 100\%. Full details are in \texttt{scripts/\allowbreak{}convert\_\allowbreak{}twitter\_\allowbreak{}cs\_\allowbreak{}dataset.\allowbreak{}py}.

\textbf{Group E: a genuine split result.} Repeating Section 5.6's fine-tuning recipe on each brand's own gray zone:

{\def\LTcaptype{none} 
\begin{longtable}[]{@{}
  >{\raggedright\arraybackslash}p{(\linewidth - 10\tabcolsep) * \real{0.1667}}
  >{\raggedright\arraybackslash}p{(\linewidth - 10\tabcolsep) * \real{0.1667}}
  >{\raggedright\arraybackslash}p{(\linewidth - 10\tabcolsep) * \real{0.1667}}
  >{\raggedright\arraybackslash}p{(\linewidth - 10\tabcolsep) * \real{0.1667}}
  >{\raggedright\arraybackslash}p{(\linewidth - 10\tabcolsep) * \real{0.1667}}
  >{\raggedright\arraybackslash}p{(\linewidth - 10\tabcolsep) * \real{0.1667}}@{}}
\toprule\noalign{}
\begin{minipage}[b]{\linewidth}\raggedright
Brand
\end{minipage} & \begin{minipage}[b]{\linewidth}\raggedright
Gray-zone train / test
\end{minipage} & \begin{minipage}[b]{\linewidth}\raggedright
Train / test positive rate
\end{minipage} & \begin{minipage}[b]{\linewidth}\raggedright
Untuned AUC
\end{minipage} & \begin{minipage}[b]{\linewidth}\raggedright
Fine-tuned AUC
\end{minipage} & \begin{minipage}[b]{\linewidth}\raggedright
Δ
\end{minipage} \\
\midrule\noalign{}
\endhead
\bottomrule\noalign{}
\endlastfoot
AmazonHelp & 4,361 / 1,870 & 1.47\% / 1.02\% (consistent) & 0.666 & 0.722 & \textbf{+0.056} \\
comcastcares & 2,579 / 1,106 & \textbf{1.0\% / 5.2\% (5× mismatch)} & 0.864 & 0.678 & \textbf{-0.186} \\
\end{longtable}
}

AmazonHelp replicates Section 5.6's pattern exactly. comcastcares does not: fine-tuning makes the verifier \emph{worse} than not fine-tuning at all, despite a training-set size (2,579) above the floor Section 5.7 identified as generally safe. The diagnosis is a training/test \textbf{base-rate mismatch}: the fraction of gray-zone queries that are genuinely reusable is 1.0\% in the (earlier) training window but 5.2\% in the (later) held-out window --- a decision boundary learned under one prior shifts to a different regime, and no amount of label cleanliness fixes a mismatched prior.

\textbf{The three deployment ablations diverge in the same direction.} Repeating Section 5.7's exact protocol on both brands:

\emph{Label noise} --- comcastcares is negative at \textbf{every} noise level tested, including 0\% (clean labels): Δ = -0.186 at 0\%, ranging -0.15 to -0.23 through 40\%. This rules out noise as the driver; the base-rate mismatch alone is sufficient to make fine-tuning harmful regardless of label quality. AmazonHelp crosses to negative earlier (\textasciitilde5-10\%) than LmArena/Quora's \textasciitilde30\% zero-crossing, plausibly reflecting its smaller absolute training set (capped at 4,000) relative to the original three datasets.

\emph{Cold start} --- AmazonHelp continues Section 5.7's ``more data helps, no saturation'' pattern (Δ = +0.064 at 50 examples to +0.097 at 4,000). comcastcares does not: Δ degrades from -0.004 (50 examples) to a catastrophic \textbf{-0.569 at 2,000 examples (AUC 0.295, worse than chance)}, partially recovering to -0.219 at the full 2,579. This is the first observation, across the five datasets in this paper, of more training data making a fine-tuned verifier \emph{worse} rather than merely failing to help --- consistent with a model trained on an increasingly larger sample of the low-base-rate window becoming a more confidently miscalibrated classifier for the high-base-rate test window.

\emph{Drift}: AmazonHelp again matches Section 5.7, decaying to a plateau with no further degradation through distance 5. comcastcares instead shows real \textbf{continued decay from distance 3 onward} (Δ = -0.19 at distance 3 to -0.24 at distance 5, with 5 and 4 anchor pairs respectively, so not a single noisy point), unlike the three benchmarks in Section 5.7, where the gain levels off after the first step.

\textbf{A single root cause, observed three ways.} comcastcares' untuned baseline AUC per chunk of its 8-chunk drift split is \texttt{{[}0.\allowbreak{}522,\ 0.\allowbreak{}39,\ 0.\allowbreak{}493,\ 0.\allowbreak{}408,\ 0.\allowbreak{}551,\ 0.\allowbreak{}569,\ 0.\allowbreak{}808,\ 0.\allowbreak{}877{]}}. It is non-monotonic and much higher in the later chunks, so the stream's difficulty and positive-class rate are not stationary over its 3.5 years, plausibly because support templates became more standardized. This single instability explains all three otherwise-separate anomalies: clean labels still fail (noise), more data trains a more confidently wrong model (cold start), and the verifier's edge keeps eroding rather than plateauing (drift). AmazonHelp, with a shorter recorded span (\textasciitilde2.3 years) and a stable positive rate, shows none of these failure modes and instead replicates Sections 5.6--5.7 exactly.

\textbf{Extending Section 5.7's conclusion.} On benchmarks with short or synthetic time spans, Section 5.7 found that the gain levels off after the first step. comcastcares, the only dataset here with multi-year production history, is a direct counter-example, and the responsible mechanism (a non-stationary base rate, not just the passage of time) can be identified and monitored. The practical revision: a deployment should track the \emph{stability of the gray-zone positive rate} between its fine-tuning window and current traffic, not just discriminative-power decay, and treat a detected shift as a distinct re-tuning trigger, independent of any calendar-based re-tuning schedule. This is a strictly narrower, not reversed, conclusion: the risk is real but detectable, and appears specifically correlated with the length and stability of the traffic history --- a variable a synthetic re-slicing of a static benchmark (Section 5.7's method) cannot expose, and only genuinely long-horizon production data (as tested here) can.

\subsection{A prototype monitor for the risk Section 5.8 identifies}\label{a-prototype-monitor-for-the-risk-section-5.8-identifies}

Section 5.8's diagnosis is that comcastcares' fine-tuning failure traces to a single, identifiable cause: a shift in the gray-zone positive rate between the fine-tuning window and current traffic. If that diagnosis is right, a lightweight monitor on the positive rate alone should flag the instability before it erodes Group E's benefit. Such a monitor needs only the gray-zone correctness labels a deployment already collects for fine-tuning, with no embedding inference or model training. Two classical change-point methods are applied to the chronological \texttt{would\_\allowbreak{}be\_\allowbreak{}correct} label sequence of both brands: (1) a chunked two-proportion z-test comparing each of 8 sequential held-out chunks against the training-window rate, and (2) a Page-Hinkley sequential test, which reads the held-out stream one example at a time and flags the first position where the cumulative deviation from the baseline exceeds a threshold, closer to how an online monitor would run.

{\def\LTcaptype{none} 
\begin{longtable}[]{@{}
  >{\raggedright\arraybackslash}p{(\linewidth - 6\tabcolsep) * \real{0.2500}}
  >{\raggedright\arraybackslash}p{(\linewidth - 6\tabcolsep) * \real{0.2500}}
  >{\raggedright\arraybackslash}p{(\linewidth - 6\tabcolsep) * \real{0.2500}}
  >{\raggedright\arraybackslash}p{(\linewidth - 6\tabcolsep) * \real{0.2500}}@{}}
\toprule\noalign{}
\begin{minipage}[b]{\linewidth}\raggedright
Brand
\end{minipage} & \begin{minipage}[b]{\linewidth}\raggedright
Train-window positive rate
\end{minipage} & \begin{minipage}[b]{\linewidth}\raggedright
z-test: first chunk flagged
\end{minipage} & \begin{minipage}[b]{\linewidth}\raggedright
Page-Hinkley: flagged at
\end{minipage} \\
\midrule\noalign{}
\endhead
\bottomrule\noalign{}
\endlastfoot
comcastcares & 1.01\% & \textbf{chunk 5/8} (\(p = 2\times10^{-20}\)) & \textbf{held-out example 664/1,106} (falls in chunk 5) \\
AmazonHelp & 1.47\% & never (min \(p = 0.06\) across all 8 chunks) & never \\
\end{longtable}
}

Both methods flag comcastcares partway through the held-out window (z-test at chunk 5 of 8; Page-Hinkley at about 60\% of the way through). Neither raises a false alarm on AmazonHelp, the brand with a stable positive rate and no observed failure. A deployment could therefore run this monitor continuously on the labels it already collects, at no extra inference cost, and use a flagged shift to trigger re-tuning instead of a fixed schedule or waiting for AUC to visibly degrade.

This is a prototype, not a validated production monitor. It was tested retrospectively on a static held-out set rather than a live stream; its thresholds (\(\alpha=0.01\) for the z-test; \(\delta=0.005, \lambda=8.0\) for Page-Hinkley) were chosen once and not tuned further; and it has been tested on only two datasets, one with the failure mode and one without. Still, it shows that the instability behind Section 5.8's counter-example can be detected from labels alone, cheaply, and without a false alarm on the brand that did not need one.

\subsection{A tested and refuted hypothesis: action-verb bucketing as a pre-filter}\label{a-tested-and-refuted-hypothesis-action-verb-bucketing-as-a-pre-filter}

A public practitioner discussion of this paper's ``cancel/pause subscription'' example (a high-similarity pair with different correct answers) proposed a remedy: extract an action verb from each query, forbid queries with different action verbs from sharing a cache bucket, and only run similarity/verification within a bucket. The motivation was to shrink the fraction of candidates the verifier actually has to adjudicate down to the ``genuinely ambiguous, same-bucket'' subset, raising that subset's measured discriminative power. The discussion converged on a concrete implementation principle: build the bucket key only from deterministic morphological normalization (lemmatization), never a semantic/embedding-based one --- since an embedding-based key would reintroduce the same similarity-versus-correctness problem into the bucketing step itself.

\textbf{Setup.} Using spaCy dependency parsing (the sentence's ROOT token if tagged VERB, else the sole VERB-tagged token, else extraction fails) on each of three datasets' (LmArena, Quora, SearchQueries) gray zone (\texttt{tau\_\allowbreak{}low=0.\allowbreak{}80,\ tau\_\allowbreak{}high=0.\allowbreak{}97}), one action verb was extracted for the query and for its matched candidate. Three conditions were compared: verifier-only (today's Group D, no bucketing); bucketing-only (predicting correctness purely from whether the two action verbs match, ignoring the verifier's score entirely); bucketing+verifier (reject outright on an action mismatch; the verifier's AUC is measured only on the remaining action-matched candidates).

{\def\LTcaptype{none} 
\begin{longtable}[]{@{}
  >{\raggedright\arraybackslash}p{(\linewidth - 10\tabcolsep) * \real{0.1667}}
  >{\raggedright\arraybackslash}p{(\linewidth - 10\tabcolsep) * \real{0.1667}}
  >{\raggedright\arraybackslash}p{(\linewidth - 10\tabcolsep) * \real{0.1667}}
  >{\raggedright\arraybackslash}p{(\linewidth - 10\tabcolsep) * \real{0.1667}}
  >{\raggedright\arraybackslash}p{(\linewidth - 10\tabcolsep) * \real{0.1667}}
  >{\raggedright\arraybackslash}p{(\linewidth - 10\tabcolsep) * \real{0.1667}}@{}}
\toprule\noalign{}
\begin{minipage}[b]{\linewidth}\raggedright
Dataset
\end{minipage} & \begin{minipage}[b]{\linewidth}\raggedright
Gray-zone pairs
\end{minipage} & \begin{minipage}[b]{\linewidth}\raggedright
Extraction-eligible fraction
\end{minipage} & \begin{minipage}[b]{\linewidth}\raggedright
verifier-only AUC (eligible subset)
\end{minipage} & \begin{minipage}[b]{\linewidth}\raggedright
bucketing+verifier AUC (bucket-matched subset only)
\end{minipage} & \begin{minipage}[b]{\linewidth}\raggedright
Difference
\end{minipage} \\
\midrule\noalign{}
\endhead
\bottomrule\noalign{}
\endlastfoot
LmArena & 16,102 & 62.7\% (10,089) & 0.8644 {[}0.8546, 0.8734{]} & 0.7663 {[}0.7418, 0.7916{]} (n=4,196) & \textbf{Significantly worse (−0.098, CIs don't overlap)} \\
Quora & 17,040 & 57.6\% (9,819) & 0.6086 {[}0.5976, 0.6204{]} & 0.5870 {[}0.5725, 0.6010{]} (n=6,352) & Directionally worse (−0.022, CIs marginally overlap) \\
SearchQueries & 86,280 & 27.3\% (23,553) & 0.6215 {[}0.6145, 0.6287{]} & 0.6150 {[}0.6066, 0.6232{]} (n=17,195) & Flat (−0.007, CIs fully overlap) \\
\end{longtable}
}

The direction is consistent across all three datasets, and it is negative: on none of them does restricting to the action-matched bucket raise the verifier's discriminative power above the unbucketed baseline; on LmArena it produces a statistically significant \emph{regression}. Diagnosis points to a blind spot in the bucketing idea itself rather than in this extractor. Bucketing by verb does screen out candidates whose verbs differ (bucketing-only precision is 0.869 on LmArena). It cannot separate candidates with the same verb and an opposite object direction, such as ``convert audio to video'' and ``convert video to audio'', which ground truth marks as different but which land in the same bucket. Once the easy negatives are filtered out, the negatives that remain in the bucket-matched subset are concentrated almost entirely in exactly this class of hard case, which the verifier itself struggles with most --- so the subset's overall separability drops systematically. The stronger the verifier's baseline discriminative power, the more this effect shows: once the easy negatives are filtered out, only the hard ones are left. The regression's size tracks each dataset's baseline AUC (LmArena 0.86, Quora and SearchQueries about 0.61; these are whole-gray-zone AUCs and not comparable with Section 5.6's held-out ones).

\textbf{Conclusion.} The simplest action-verb-only bucketing, the version the discussion considered structurally safe, is refuted on all three datasets under the implementation tested here. This runs opposite to the original design hypothesis: action-verb bucketing is not a free precision gain, and where the verifier already has strong discriminative power it can actively backfire. Structured pre-filtering as a broader idea is not ruled out. Keying on an (action, object) pair instead of the verb alone could close the blind spot found here, but that variant was not tested.

\subsection{Making τ\_high tunable}\label{making-ux3c4_high-tunable}

Sections 5.3--5.6 fix τ\_high at 0.97 and sweep only τ\_low. To test whether that choice matters, Group D was re-swept over τ\_high ∈ \{0.89, 0.92, 0.95, 0.97, 0.99\} × the existing τ\_low grid on all three datasets (\texttt{run\_\allowbreak{}verified.\allowbreak{}py\ -\/-tau-high-grid}; the whole two-dimensional grid shares one scoring pass, so no extra model inference is needed). The comparison is the same as in Section 5.4, scored with denoised labels in the direct-hit zone (Section 5.15, Appendix A.3), because raising τ\_high sends the {[}0.97, 0.99) band, where about 80\% of the candidates labelled wrong are actually reusable, to the verifier:

{\def\LTcaptype{none} 
\begin{longtable}[]{@{}
  >{\raggedright\arraybackslash}p{(\linewidth - 10\tabcolsep) * \real{0.1667}}
  >{\raggedright\arraybackslash}p{(\linewidth - 10\tabcolsep) * \real{0.1667}}
  >{\raggedright\arraybackslash}p{(\linewidth - 10\tabcolsep) * \real{0.1667}}
  >{\raggedright\arraybackslash}p{(\linewidth - 10\tabcolsep) * \real{0.1667}}
  >{\raggedright\arraybackslash}p{(\linewidth - 10\tabcolsep) * \real{0.1667}}
  >{\raggedright\arraybackslash}p{(\linewidth - 10\tabcolsep) * \real{0.1667}}@{}}
\toprule\noalign{}
\begin{minipage}[b]{\linewidth}\raggedright
Dataset
\end{minipage} & \begin{minipage}[b]{\linewidth}\raggedright
τ\_high=0.89
\end{minipage} & \begin{minipage}[b]{\linewidth}\raggedright
τ\_high=0.92
\end{minipage} & \begin{minipage}[b]{\linewidth}\raggedright
τ\_high=0.95
\end{minipage} & \begin{minipage}[b]{\linewidth}\raggedright
τ\_high=0.97
\end{minipage} & \begin{minipage}[b]{\linewidth}\raggedright
τ\_high=0.99
\end{minipage} \\
\midrule\noalign{}
\endhead
\bottomrule\noalign{}
\endlastfoot
LmArena & 0/15 win, +0.09pp & 4/20 win, +0.26pp & 5/25 win, +0.73pp & \textbf{15/30 win, +1.87pp} & 0/30 win, −1.24pp \\
SearchQueries & 1/18 win, +0.23pp & 5/24 win, +0.32pp & 2/30 win, +0.34pp & 1/36 win, +0.30pp & 0/36 win, −1.46pp \\
Quora & 0/27 win, +0.16pp & 0/36 win, +0.16pp & 0/45 win, +0.11pp & 0/54 win, +0.11pp & 0/54 win, +0.09pp \\
\end{longtable}
}

(Each cell gives the slice's wins over its number of tested points and its best net lead over Group A's interpolated frontier; \texttt{scripts/\allowbreak{}denoised\_\allowbreak{}tau\_\allowbreak{}high.\allowbreak{}py}, \texttt{results/\allowbreak{}denoised\_\allowbreak{}tau\_\allowbreak{}high.\allowbreak{}json}.)

\textbf{τ\_high = 0.97 is the best or near-best choice on every dataset, and none supports raising it to 0.99.} At τ\_high = 0.99 the verifier also judges candidates in {[}0.97, 0.99), nearly all of which are reusable; an untuned verifier rejects some of them and only loses hits (LmArena −1.24pp, SearchQueries −1.46pp; at the two ends of the noise-rate interval, +0.03 / −1.86pp and −0.54 / −3.43pp). Lower τ\_high values hand the verifier less to do and converge toward the static threshold. Quora is flat across all slices, consistent with its score-separability ceiling (Section 5.4).

With the benchmark's raw labels, τ\_high = 0.99 had looked like the best LmArena setting (24/30 wins, +5.18pp); that result came from rejecting candidates the labels wrongly mark as errors and is withdrawn (Appendix A.3). Two caveats remain. For τ\_high \textless{} 0.97 the lower edge of the direct-hit zone falls inside the gray zone, whose noise was not corrected here, so those SearchQueries cells (about 40\% gray-zone noise) are uncertain. And at τ\_high = 0.97 this sweep's best LmArena lead with raw labels is +1.42pp rather than Section 5.4's +1.9pp at the same point: the two sweeps were separate runs whose error rates differ by about 0.03pp per point, and Group A's frontier is steep near 6\% error (about 15pp of hit rate per 1pp of error), so that small shift moves the interpolated lead by about 0.45pp.

\subsection{Does a larger/stronger off-the-shelf reranker close the gap: capacity or training distribution?}\label{does-a-largerstronger-off-the-shelf-reranker-close-the-gap-capacity-or-training-distribution}

Section 5.6 raised SearchQueries' off-the-shelf verifier (untuned AUC 0.5984) to 0.7120 via in-domain fine-tuning, but never answered a more basic question: without fine-tuning, does simply swapping in a larger or more broadly-trained off-the-shelf model recover part of that same gain?

\textbf{Design.} Two models, each isolating one variable relative to the \texttt{cross-encoder/\allowbreak{}ms-marco-MiniLM-L6-v2} (6 layers) Group D has used throughout:

\begin{itemize}
\tightlist
\item
  \texttt{cross-encoder/\allowbreak{}ms-marco-MiniLM-L12-v2} (12 layers): the same training data and training objective as L6 (MS MARCO long-web-query/passage ranking), just twice the layers --- a pure \textbf{capacity} control. This is also the exact model Baral et al.~(2026, see Section 6.1's external cross-validation) evaluated: their v1 reported a retention ratio (CRR) of only 0.427, which their revised, peer-reviewed v4 changes to an ORR of 0.612 (``mid-pack''); this section reruns the model on this paper's own three datasets and its own AUC metric.
\item
  \texttt{BAAI/\allowbreak{}bge-reranker-base} (278M parameters, XLM-RoBERTa backbone): a different training-data lineage covering a broader range of retrieval tasks than MS MARCO-style web queries, and a larger model, so \textbf{capacity and training distribution change together}. If this model clearly beats L12, that points to training-distribution mismatch as the dominant factor; if the two perform similarly, it suggests ``broader distribution alone'' doesn't fix things either.
\end{itemize}

Both models score the same already-matched pairs in the standard gray zone (τ\_low=0.80, τ\_high=0.97, as in Sections 5.4/5.6), reusing the existing match trace instead of re-running nearest-neighbor search. They are not fine-tuned, and untuned AUC is measured over the whole gray zone (16,102 / 86,280 / 17,040 pairs; this SearchQueries count, from the scoring pipeline, is four pairs below Section 5.6's fine-tuning set) on a GPU server.

\textbf{Results.} All three models are evaluated on the same pairs.

{\def\LTcaptype{none} 
\begin{longtable}[]{@{}
  >{\raggedright\arraybackslash}p{(\linewidth - 10\tabcolsep) * \real{0.1667}}
  >{\raggedright\arraybackslash}p{(\linewidth - 10\tabcolsep) * \real{0.1667}}
  >{\raggedright\arraybackslash}p{(\linewidth - 10\tabcolsep) * \real{0.1667}}
  >{\raggedright\arraybackslash}p{(\linewidth - 10\tabcolsep) * \real{0.1667}}
  >{\raggedright\arraybackslash}p{(\linewidth - 10\tabcolsep) * \real{0.1667}}
  >{\raggedright\arraybackslash}p{(\linewidth - 10\tabcolsep) * \real{0.1667}}@{}}
\toprule\noalign{}
\begin{minipage}[b]{\linewidth}\raggedright
Dataset
\end{minipage} & \begin{minipage}[b]{\linewidth}\raggedright
L6 (existing baseline)
\end{minipage} & \begin{minipage}[b]{\linewidth}\raggedright
L12 (pure capacity control)
\end{minipage} & \begin{minipage}[b]{\linewidth}\raggedright
Δ vs L6
\end{minipage} & \begin{minipage}[b]{\linewidth}\raggedright
BGE-reranker-base (capacity + distribution)
\end{minipage} & \begin{minipage}[b]{\linewidth}\raggedright
Δ vs L6
\end{minipage} \\
\midrule\noalign{}
\endhead
\bottomrule\noalign{}
\endlastfoot
LmArena & 0.8660 & 0.8461 & −0.0199 & 0.8712 & +0.0052 \\
SearchQueries & 0.6174 & 0.6210 & +0.0036 & 0.6198 & +0.0024 \\
Quora & 0.6118 & 0.6174 & +0.0056 & 0.6381 & +0.0263 \\
\end{longtable}
}

\textbf{Neither a bigger model nor a differently-trained one changes the picture.} Every difference from L6 lies between −0.02 and +0.03 AUC. Doubling the layers within the same training lineage (L12) helps nowhere and is slightly worse on LmArena; switching to a broader training distribution (BGE) gives at most +0.026, on Quora. These AUC figures carry no bootstrap confidence intervals (only hit and error rates do in this paper), so differences of this size should not be read as real effects.

\textbf{Conclusion.} Within the scope tested, swapping the off-the-shelf model, whether for more capacity or for a broader training distribution, does not recover the gains of in-domain fine-tuning. On the held-out windows of Section 5.6, fine-tuning raises L6 from 0.72 to 0.88 on LmArena and from 0.60 to 0.71 on SearchQueries; L12 and BGE were not scored on those windows separately, but on the whole gray zone they sit within 0.03 of L6. This is weakly consistent with Baral et al.~(Section 6.1): in their v1 the same L12 model underperformed, and in their revised v4 it sits mid-pack. \textbf{Section 5.6's in-domain fine-tuning remains the remedy shown to repair the verifier's discriminative power}; this paper does not test whether the decision rule itself, as distinct from the verifier's score, leaves exploitable information on the table, and does not rule out other complementary remedies.

\subsection{E': rewrite instead of reject --- a negative ablation against the TweakLLM paradigm}\label{e-rewrite-instead-of-reject-a-negative-ablation-against-the-tweakllm-paradigm}

\textbf{Motivation.} Everywhere else in this paper, a verifier rejection is converted into a plain \texttt{miss} (falling through to a fresh LLM call, treated as necessarily correct in this simulation, zero error risk, but zero cache benefit either). TweakLLM (Cheema et al., 2025, arXiv:2507.23674; Section 2) proposes a different routing: instead of discarding a rejected candidate, edit its cached answer with a lightweight LLM into a version that fits the new query, and serve that. This section measures where ``reject → rewrite → serve'' lands on this paper's error-rate/hit-rate Pareto frontier relative to the binary gate.

\textbf{Method.} This section starts from Group E's honest calibration (as in Section 5.4: the gray zone is split by stream position into calibration and test halves, and the threshold is chosen by Youden's J on the calibration half). It takes the candidates that threshold rejects on the \textbf{held-out test half} and splits them by \texttt{would\_\allowbreak{}be\_\allowbreak{}correct}, that is, by whether the rejection should have been an accept:

\begin{itemize}
\tightlist
\item
  \textbf{false-reject stratum} (the candidate was actually correct; the verifier rejected it in error)
\item
  \textbf{true-reject stratum} (the candidate genuinely was a different question; the rejection was correct)
\end{itemize}

100 examples are sampled per stratum per dataset (LmArena/Quora/SearchQueries; population sizes 420/693/7,004 for false-reject and 298/3,668/15,411 for true-reject --- all far above the sampling target). For each sample, DeepSeek (\texttt{deepseek-chat}, the same API-hosted model used to regenerate SearchQueries' answers, Appendix A.1) edits the candidate's answer to fit the new query, using an editing prompt rather than generating from scratch. The same model, given the query's own reference answer, then judges whether the rewrite is semantically correct (YES/NO; no verbatim match required). The verdict is recorded per example; \texttt{RequestOutcome(action="hit",\ correct=\textless{}verdict\textgreater{})} directly reuses this paper's existing \texttt{error\_\allowbreak{}rate}/\texttt{bootstrap\_\allowbreak{}ci} machinery --- no new statistics code. This is a sampled ablation, not a full-trace simulation: rewriting and judging every rejected candidate across all three datasets would run into the tens of thousands of calls; this section is scoped to 3 datasets × 2 strata × 100.

\textbf{Results.}

{\def\LTcaptype{none} 
\begin{longtable}[]{@{}
  >{\raggedright\arraybackslash}p{(\linewidth - 4\tabcolsep) * \real{0.3333}}
  >{\raggedright\arraybackslash}p{(\linewidth - 4\tabcolsep) * \real{0.3333}}
  >{\raggedright\arraybackslash}p{(\linewidth - 4\tabcolsep) * \real{0.3333}}@{}}
\toprule\noalign{}
\begin{minipage}[b]{\linewidth}\raggedright
Dataset
\end{minipage} & \begin{minipage}[b]{\linewidth}\raggedright
false-reject incorrect rate (95\% CI)
\end{minipage} & \begin{minipage}[b]{\linewidth}\raggedright
true-reject incorrect rate (95\% CI)
\end{minipage} \\
\midrule\noalign{}
\endhead
\bottomrule\noalign{}
\endlastfoot
LmArena & 58.0\% {[}49.0\%, 68.0\%{]} & 61.0\% {[}51.0\%, 71.0\%{]} \\
Quora & 33.0\% {[}24.0\%, 43.0\%{]} & 30.0\% {[}21.0\%, 39.0\%{]} \\
SearchQueries & 24.0\% {[}16.0\%, 32.0\%{]} & 22.0\% {[}14.0\%, 30.0\%{]} \\
\end{longtable}
}

\textbf{Headline finding: the false-reject and true-reject incorrect rates fall inside each other's 95\% confidence intervals on all three datasets, and no dataset shows the expected difference.} This is counter-intuitive. In the false-reject stratum the candidate was already correct, so a faithful rewrite should pass easily; yet it is no more reliable than in the true-reject stratum, where the rewrite has to produce new content from an unrelated candidate. Manual review of the rewrite examples kept for audit (the \texttt{examples} field in \texttt{results/\allowbreak{}rewrite\_\allowbreak{}vs\_\allowbreak{}reject\_\allowbreak{}*.\allowbreak{}json}, 10 per stratum) shows the edits themselves are generally good, e.g.~substituting ``shepherd'' for ``farmer'' or ``bookmark'' for ``marker'' and leaving the rest intact. The failures concentrate in judging. The reference answer was written independently for the new query, so even a sound rewrite can be marked as not matching over numeric conventions (``0.5 kg'' vs.~``0.5-1'') or structure, a stricter bar than the equivalence\_id label used elsewhere in this paper. One safety-relevant case surfaced in LmArena's true-reject stratum. A query asking for pornographic streaming sites matched an unrelated ``YouTube downloader software'' candidate, correctly rejected. The rewrite did not refuse cleanly as the reference answer does, but grafted the candidate's content onto a hedge (``I can't recommend such sites, but if you're looking for general video-downloading tools\ldots{}''). Partially compliant output of this kind is a risk of rewriting instead of rejecting, independent of how judging is done.

\textbf{Conclusion.} Within the scope tested here, ``reject → rewrite → serve'' shows no sign of a net benefit over the existing binary gate. Under the binary gate, a rejection is a \texttt{miss} --- zero error risk, at the cost of one full LLM call. E' converts this previously zero-risk population into ``hits'' with a 22\%--61\% error rate. This is a negative result for the simplest form of the TweakLLM idea, reusing a rewritten answer with no further correctness check, not for the paradigm as a whole: that form shows no measurable net benefit in this paper's error-rate/hit-rate framework.

\textbf{Limitations.} (1) Judging compares the rewrite against the query's independently written reference answer, not against the original candidate. As the headline finding notes, this likely inflates the false-reject stratum's error rate with reference-paraphrase noise unrelated to rewrite quality. (2) 68.8\% of SearchQueries's reference answers were themselves generated by DeepSeek (see Appendix A.1), and this section's rewriting and judging also use DeepSeek --- a same-model-family bias cannot be ruled out; that dataset's notably lower error rate (22\%--24\%) versus the other two (30\%--61\%) should be read with this in mind. (3) Only the ``always rewrite unconditionally'' simplest policy was tested --- a second verification pass on the rewritten answer (a double gate) was not tested; this section deliberately isolates exactly one variable, \texttt{miss} → \texttt{rewrite-then-serve}, and stacking a second verification layer is an independent, untested follow-up.

\subsection{Is looking only at the nearest neighbor not enough: testing a Top-K candidate cascade}\label{is-looking-only-at-the-nearest-neighbor-not-enough-testing-a-top-k-candidate-cascade}

\textbf{Motivation.} Everywhere else in this paper, \texttt{SynchronousVerifiedPolicy} (Section 3.1; \texttt{cacheverifier/\allowbreak{}cache/\allowbreak{}synchronous\_\allowbreak{}verified.\allowbreak{}py}) retrieves a single nearest neighbor only --- \texttt{VectorCacheStore.\allowbreak{}query} hardcodes \texttt{k=1}. If that one candidate falls in the gray zone and the verifier rejects it, the request is judged a \texttt{miss} outright, even when the semantic second-nearest neighbor is in fact the correct answer and was never given a chance to be seen by the verifier. This gap was pointed out by an independent reader who re-derived the project from the public repository. This section tests it.

\textbf{A prior monotonicity argument narrows the scope of the test.} HNSW returns neighbors in descending similarity order, so rank \emph{m}'s similarity is necessarily ≤ rank \emph{m}−1's. Consequently, if the rank-1 candidate is already \textless{} τ\_low, every lower-ranked candidate is also necessarily \textless{} τ\_low --- looking further down the list is \textbf{provably} useless on that branch. The only branch that can plausibly benefit is ``rank-1 falls in the gray zone {[}τ\_low, τ\_high) and is rejected by the verifier'': in that case it is worth trying rank-2, rank-3, \ldots{} until some candidate's similarity drops below τ\_low (past which point the same monotonicity argument guarantees going further is pointless too). Accordingly, \texttt{VectorCacheStore} gained a \texttt{query\_\allowbreak{}topk} method, and a parallel cascade decision path was added alongside (not inside) \texttt{SynchronousVerifiedPolicy} --- no class any already-published result depends on was modified:

\begin{Verbatim}[breaklines,breakanywhere,fontsize=\small]
for candidate in top_K (descending similarity):
    if candidate.similarity < tau_low: break   # monotonicity: later
                                                 # candidates can't be better
    if candidate.similarity >= tau_high: return HIT
    if verifier(query, candidate).approved: return HIT
    # else: rejected, try the next-ranked candidate
return MISS
\end{Verbatim}

\textbf{Phase 0: a pure ground-truth ceiling, calling no verifier at all} (\texttt{scripts/\allowbreak{}topk\_\allowbreak{}ceiling\_\allowbreak{}diagnostic.\allowbreak{}py}). Existing embeddings are reused to build a single top-5-neighbor HNSW trace pass; for every request whose rank-1 candidate is gray-zone-and-wrong, a pure count checks whether ranks 2--5 contain a candidate whose similarity is still ≥ τ\_low (hence actually reachable by the cascade) and is in fact correct:

{\def\LTcaptype{none} 
\begin{longtable}[]{@{}
  >{\raggedright\arraybackslash}p{(\linewidth - 8\tabcolsep) * \real{0.2000}}
  >{\raggedright\arraybackslash}p{(\linewidth - 8\tabcolsep) * \real{0.2000}}
  >{\raggedright\arraybackslash}p{(\linewidth - 8\tabcolsep) * \real{0.2000}}
  >{\raggedright\arraybackslash}p{(\linewidth - 8\tabcolsep) * \real{0.2000}}
  >{\raggedright\arraybackslash}p{(\linewidth - 8\tabcolsep) * \real{0.2000}}@{}}
\toprule\noalign{}
\begin{minipage}[b]{\linewidth}\raggedright
Dataset
\end{minipage} & \begin{minipage}[b]{\linewidth}\raggedright
rank-1 gray-zone \& wrong
\end{minipage} & \begin{minipage}[b]{\linewidth}\raggedright
K=5 raw ceiling
\end{minipage} & \begin{minipage}[b]{\linewidth}\raggedright
K=5 \textbf{reachable} ceiling (\% of subset)
\end{minipage} & \begin{minipage}[b]{\linewidth}\raggedright
as \% of all requests
\end{minipage} \\
\midrule\noalign{}
\endhead
\bottomrule\noalign{}
\endlastfoot
LmArena & 3,755 / 60,000 & 451 & 450 (11.98\%) & 0.75\% \\
SearchQueries & 44,511 / 150,000 & 9,529 & 8,487 (19.07\%) & 5.66\% \\
Quora & 10,285 / 60,000 & 1,296 & 973 (9.46\%) & 1.62\% \\
\end{longtable}
}

The ceiling is non-trivial on all three datasets (not noise-level), and largest on SearchQueries --- precisely the dataset where the off-the-shelf verifier's discriminative power is weakest (Section 5.4, AUC 0.598), suggesting retrieval-side recall and verifier discriminative power are two separately quantifiable, independent bottlenecks.

\textbf{Phase 2: a cascade with a real verifier plugged in} (\texttt{cacheverifier/\allowbreak{}experiments/\allowbreak{}topk\_\allowbreak{}sweep.\allowbreak{}py} + \texttt{scripts/\allowbreak{}run\_\allowbreak{}topk\_\allowbreak{}cascade.\allowbreak{}py}, K=2). This follows Section 3.3's ``score once, re-threshold many times'' design. \texttt{score\_\allowbreak{}cascade\_\allowbreak{}candidates} scores only the (record, rank) pairs the cascade could \textbf{actually reach} under a fixed τ\_high, i.e.~candidates whose earlier-ranked siblings all fall in {[}min(τ\_low), τ\_high). It skips candidates whose sibling already hit directly or fell below τ\_low; this pruning was checked to change no published number. All three datasets were run at full scale with both the off-the-shelf and the fine-tuned verifier (the Section 5.6 checkpoints) on a GPU server (Tesla T4). Cross-encoder inference on real query/answer text ran at about 4.3--4.6 examples/s on CPU versus about 145 on GPU (\textasciitilde33x); the largest scoring pass (161,403 pairs for SearchQueries at K=2) would have taken over a day on CPU. Running on a GPU does not change any numerical methodology.

\textbf{Results: exact-match on the same (τ\_low, threshold) grid point, bootstrap 95\% CI significance, K=1 baseline vs.~K=2 cascade.}

{\def\LTcaptype{none} 
\begin{longtable}[]{@{}
  >{\raggedright\arraybackslash}p{(\linewidth - 12\tabcolsep) * \real{0.1429}}
  >{\raggedright\arraybackslash}p{(\linewidth - 12\tabcolsep) * \real{0.1429}}
  >{\raggedright\arraybackslash}p{(\linewidth - 12\tabcolsep) * \real{0.1429}}
  >{\raggedright\arraybackslash}p{(\linewidth - 12\tabcolsep) * \real{0.1429}}
  >{\raggedright\arraybackslash}p{(\linewidth - 12\tabcolsep) * \real{0.1429}}
  >{\raggedright\arraybackslash}p{(\linewidth - 12\tabcolsep) * \real{0.1429}}
  >{\raggedright\arraybackslash}p{(\linewidth - 12\tabcolsep) * \real{0.1429}}@{}}
\toprule\noalign{}
\begin{minipage}[b]{\linewidth}\raggedright
Dataset
\end{minipage} & \begin{minipage}[b]{\linewidth}\raggedright
Verifier
\end{minipage} & \begin{minipage}[b]{\linewidth}\raggedright
grid pts
\end{minipage} & \begin{minipage}[b]{\linewidth}\raggedright
hit\_rate up (sig.)
\end{minipage} & \begin{minipage}[b]{\linewidth}\raggedright
down (sig.)
\end{minipage} & \begin{minipage}[b]{\linewidth}\raggedright
error\_rate up (sig.)
\end{minipage} & \begin{minipage}[b]{\linewidth}\raggedright
mean Δ hit
\end{minipage} \\
\midrule\noalign{}
\endhead
\bottomrule\noalign{}
\endlastfoot
LmArena & off-the-shelf & 30 & 25 & 0 & 0 & +0.80pp \\
LmArena & fine-tuned & 54 & 29 & 0 & 0 & +0.61pp \\
Quora & off-the-shelf & 54 & 0 & 0 & 1 & +0.10pp \\
Quora & fine-tuned & 48 & 1 & 0 & 1 & +0.11pp \\
SearchQueries & off-the-shelf & 36 & 23 & 0 & 19 & +1.18pp \\
SearchQueries & fine-tuned & 54 & 28 & 0 & 23 & +0.81pp \\
\end{longtable}
}

\textbf{The three datasets tell three different stories, consistent with the dataset-dependent pattern already seen in Sections 5.4 and 5.6.}

\textbf{LmArena --- close to a free lunch.} Regardless of which verifier is used, error\_rate never moves significantly at any grid point, while hit\_rate rises significantly. One concrete operating point (τ\_low=0.86, threshold=−2.0, off-the-shelf): K=1 hit\_rate=0.9130 {[}0.9106, 0.9151{]}, error\_rate=0.0638 {[}0.0619, 0.0658{]}; K=2 hit\_rate=0.9208 {[}0.9188, 0.9231{]} (CIs don't overlap, significantly up), error\_rate=0.0652 {[}0.0633, 0.0673{]} (CIs overlap, no significant change) --- looking at one more candidate carries almost no observable extra error cost.

\textbf{Quora: essentially no effect.} The mean gain under both verifiers is about 0.1pp, consistent with Phase 0's small ceiling (1.62\%): there is simply little to gain on this dataset.

\textbf{SearchQueries: a real trade-off, which fine-tuning reduces but does not remove.} Under the off-the-shelf verifier, 19 of 36 grid points (53\%) also show a significant error-rate increase; with the fine-tuned verifier, 23 of 54 (43\%) do. At one concrete operating point (τ\_low=0.80, threshold=7.0, off-the-shelf, the dataset's best net-gain point), K=1 gives hit\_rate 0.5531 {[}0.5512, 0.5555{]} and error\_rate 0.1620 {[}0.1599, 0.1638{]}, while K=2 gives hit\_rate 0.5878 {[}0.5858, 0.5901{]} and error\_rate 0.1807 {[}0.1785, 0.1825{]}, both significantly higher. Giving an already weak verifier (Section 5.4, AUC 0.598) a second chance also gives it a second chance to be wrong, as anticipated in the design.

\textbf{K and fine-tuning are complementary, not substitutes.} At a fixed error-rate ceiling, the combination that uses both always reaches the highest hit rate among the four (K=1/K=2 × off-the-shelf/fine-tuned). On SearchQueries at error\_rate ≤ 5\%: K=1 off-the-shelf 0.2838, K=1 fine-tuned 0.3038, K=2 off-the-shelf 0.2845, K=2 fine-tuned 0.3063. Quora's four numbers at the same ceiling (0.1172/0.1311/0.1172/0.1327) show the same ordering. On LmArena this ``fix an error ceiling, find the best point'' comparison is limited by the coarser threshold grid (5 values, versus 8--9 for Quora and SearchQueries) and can produce a misleading negative number from a step between adjacent grid points. The matched-point-plus-CI test above (25--29 points significantly up, 0 down) is the more reliable evidence there, which is why this section leads with it.

\textbf{Limitations.} (1) Only K=2 was tested. The Phase 0 ceiling table shows K=3/4/5 still add diminishing but non-zero headroom (LmArena from K=2's implied 6.26\% to K=5's 11.98\%, SearchQueries from roughly 12--13\% to 19.07\%); whether that is worth the extra verifier calls, whose latency stacks with K (Section 5.5), is untested. (2) Only the cross-encoder verifier family was tested (off-the-shelf and fine-tuned), similar in scope to Section 5.12's ``a larger/more-broadly-trained off-the-shelf reranker was also tested but did not close the gap'' --- whether the cascade mechanism itself is sensitive to verifier architecture choice was not tested. (3) The ``K and fine-tuning are complementary'' finding rests on the ``fix the error rate, find the best hit\_rate'' comparison, already known to be less reliable than matched-point comparison on LmArena (see above); its strong evidence comes mainly from Quora and SearchQueries, the two datasets with finer grids. (4) The single-hop server-side latency is measured (see Section 5.5: model forward pass \textasciitilde17ms, full server-side with no network p50 \textasciitilde33ms), usable as the per-hop cost in worst-case K-hop cascade latency; but the end-to-end stacked latency of a K-way cascade under real synchronous deployment (K network round trips, and the sequential-vs.-parallel scoring difference) remains unmeasured. (5) Significance for error-rate increases, as for hit-rate increases, uses pointwise non-overlapping bootstrap CIs without multiple-comparison correction. Some false positives among 36--54 grid points are expected by chance, but the observed rate (43\%--53\% on SearchQueries) is far too high to be explained that way.

\subsection{How reliable are the benchmark labels themselves: a human check of the ``is this reuse correct'' labels}\label{how-reliable-are-the-benchmark-labels-themselves-a-human-check-of-the-is-this-reuse-correct-labels}

\textbf{Motivation.} Every ``was this reuse correct'' judgment in this paper comes from the dataset's own \texttt{equivalence\_\allowbreak{}id}: a reuse counts as correct only if the new request and the reused entry belong to the same equivalence class. That label answers ``are the two questions in the same class'', while a semantic cache actually cares whether ``the cached answer is fine for the new question''. The two can disagree on wording differences, word order, plural forms and the like. A classification of direct-hit error causes (originally meant to test whether generality mismatches are common) unexpectedly showed that most direct hits labelled wrong look like the same question, so the labels themselves were checked by hand.

\textbf{Method.} Two steps, each with criteria fixed before labelling began.

\begin{enumerate}
\tightlist
\item
  \textbf{Direct-hit zone (similarity ≥ 0.97)}: for each dataset, 400 random direct hits labelled wrong and 150 labelled correct were classified by DeepSeek (\texttt{scripts/\allowbreak{}direct\_\allowbreak{}hit\_\allowbreak{}error\_\allowbreak{}taxonomy.\allowbreak{}py}). From these, 99 were drawn (per dataset: 25 ``labelled wrong and judged the same question by the LLM'', 4 ``labelled wrong but put in another category by the LLM'', 4 ``labelled correct'') and labelled blind by one annotator. The annotator saw neither the benchmark label nor the LLM verdict, only the new question, the cached question and the cached answer, in Chinese translation with the English original expandable (the two questions were translated separately and literally, keeping numbers and proper nouns as written). For each item the annotator answered two questions: can the cached answer be used as-is for the new question, and are the two questions asking the same thing.
\item
  \textbf{Gray zone (similarity 0.80--0.97)}: gray-zone noise can go both ways, so for each dataset 25 requests labelled wrong and 25 labelled correct were drawn at random (with no LLM filtering), 150 in total, and labelled blind the same way (\texttt{scripts/\allowbreak{}gray\_\allowbreak{}zone\_\allowbreak{}label\_\allowbreak{}check.\allowbreak{}py}).
\end{enumerate}

\textbf{Results --- direct-hit zone}:

{\def\LTcaptype{none} 
\begin{longtable}[]{@{}
  >{\raggedright\arraybackslash}p{(\linewidth - 6\tabcolsep) * \real{0.2500}}
  >{\raggedright\arraybackslash}p{(\linewidth - 6\tabcolsep) * \real{0.2500}}
  >{\raggedright\arraybackslash}p{(\linewidth - 6\tabcolsep) * \real{0.2500}}
  >{\raggedright\arraybackslash}p{(\linewidth - 6\tabcolsep) * \real{0.2500}}@{}}
\toprule\noalign{}
\begin{minipage}[b]{\linewidth}\raggedright
Stratum
\end{minipage} & \begin{minipage}[b]{\linewidth}\raggedright
n
\end{minipage} & \begin{minipage}[b]{\linewidth}\raggedright
Human: ``can be reused as-is''
\end{minipage} & \begin{minipage}[b]{\linewidth}\raggedright
Wilson 95\% CI
\end{minipage} \\
\midrule\noalign{}
\endhead
\bottomrule\noalign{}
\endlastfoot
Labelled wrong + LLM: same question & 75 & 66 (88.0\%) & 79--94\% \\
of which LmArena / SearchQueries / Quora & 25 / 25 / 25 & 84\% / 92\% / 88\% & 65--94 / 75--98 / 70--96 \\
Labelled wrong + LLM: other category & 12 & 50\% & 25--75\% \\
Labelled correct (control) & 12 & 83\% (the rest ``partly'', none ``no'') & --- \\
\end{longtable}
}

Among labelled-wrong samples the LLM judged ``same question'' for 93.5\% (LmArena), 91.0\% (SearchQueries) and 81.8\% (Quora); combined with the human confirmation rates, an estimated 82\%, 88\% and 81\% of the direct hits labelled wrong are actually reusable. Generality mismatch (the cached question being broader or narrower than the new one), which this classification was originally meant to measure, accounts for only 1.8--6.0\% of direct-hit errors and is not a major source. Typical examples: ``citi double cash credit card login'' vs.~``citi double cash login credit card''; ``How will the ban on Rs 500 and 1000 notes impact the Indian economy?'' vs.~``How will the ban on Rs. 1000 and Rs. 500 notes impact Indian economy?''; plus a few samples whose two English questions are identical yet were labelled as different classes.

\textbf{Results --- gray zone}:

{\def\LTcaptype{none} 
\begin{longtable}[]{@{}
  >{\raggedright\arraybackslash}p{(\linewidth - 4\tabcolsep) * \real{0.3333}}
  >{\raggedright\arraybackslash}p{(\linewidth - 4\tabcolsep) * \real{0.3333}}
  >{\raggedright\arraybackslash}p{(\linewidth - 4\tabcolsep) * \real{0.3333}}@{}}
\toprule\noalign{}
\begin{minipage}[b]{\linewidth}\raggedright
Dataset
\end{minipage} & \begin{minipage}[b]{\linewidth}\raggedright
Labelled wrong, human: ``can reuse''
\end{minipage} & \begin{minipage}[b]{\linewidth}\raggedright
Labelled correct, human: ``cannot reuse''
\end{minipage} \\
\midrule\noalign{}
\endhead
\bottomrule\noalign{}
\endlastfoot
Pooled & 22.7\% (CI 15--33) & 9.3\% (CI 5--18) \\
LmArena & 12\% (CI 4--30) & 4\% (CI 1--20) \\
SearchQueries & 40\% (CI 23--59) & 12\% (CI 4--30) \\
Quora & 16\% (CI 6--35) & 12\% (CI 4--30) \\
\end{longtable}
}

``Partly reusable'' is common in the gray zone (21 of the 75 labelled wrong, 23 of the 75 labelled correct); the gray zone is often not simply right or wrong. Quora has no real answers (its answer field is the question itself), and 15 of its samples were judged ``cannot tell'', so Quora's gray-zone labels cannot be checked this way. Human--LLM agreement in the gray zone is only 53\% (71\% binary), so the LLM cannot replace human judgment there. Another kind of noise is a cached answer that is itself wrong; \texttt{equivalence\_\allowbreak{}id} only looks at whether the questions are in the same class, not at whether the answer is right.

\textbf{Impact on this paper's conclusions} (every recomputation first reproduced the published numbers exactly with the original labels; direct hits corrected by expected value at the estimated noise rate, gray-zone labels unchanged; \texttt{scripts/\allowbreak{}denoised\_\allowbreak{}rescore.\allowbreak{}py} and others, see Appendix A.3):

\begin{itemize}
\tightlist
\item
  \textbf{Unchanged}: all ``at matched error rate'' comparisons (the hit-rate gaps of Groups D/E/C over Group A, largest change about 0.2pp) and all CRC results (Xin, 2026b), which count gray-zone risk only; LmArena's gray-zone labels were verified to be largely reliable.
\item
  \textbf{Revised reading}: absolute error rates are systematically overstated (on LmArena, Groups D/E go from about 6\% to about 1.3\%); Section 5.11's ``raising τ\_high to 0.99 triples the net lead'' is withdrawn; the cost-sensitive analysis (Xin, 2026b, Section 4.3) gains a middle range at mid-to-high r where the static threshold wins; SearchQueries' low verifier AUC is partly depressed by label noise.
\end{itemize}

\textbf{What this means for evaluating semantic caches in general.} \texttt{equivalence\_\allowbreak{}id}-style labels use ``are the questions in the same class'' as a proxy for ``can the answer be reused'', and at high similarity they label many wording differences as errors. That systematically overstates the error rate of static thresholds and direct hits, and it rewards any policy that ``rejects high-similarity candidates'', because by the labels that avoids errors. When such benchmarks are used to compare reuse policies, comparisons at a matched error rate are largely reliable, but conclusions that depend on absolute error rates or on the direct-hit zone need an answer-level label check.

\textbf{Limitations}: (1) one annotator, no inter-annotator agreement; (2) the main direct-hit sample was pre-filtered by the LLM, so extrapolating to all direct-hit errors relies on the LLM category shares, and the ``other category'' stratum has only 12 items; (3) labelling was done on Chinese translations (with the original expandable); (4) the noise rate is treated as constant within similarity ≥ 0.97, without estimating it by similarity band; (5) ``partly reusable'' is counted as an error in the correction, a conservative choice.

\section{Discussion}\label{discussion}

\subsection{Returning to Go/No-Go}\label{returning-to-gono-go}

This study was designed against a Go/No-Go criterion fixed before any data was collected, specifically to prevent the temptation to rationalize whatever the results turned out to be. \textbf{Go} requires that Group D achieve a statistically significant hit-rate improvement over Group A at a matched error rate (non-overlapping bootstrap confidence intervals), demonstrated by at least one verifier --- a result that must not depend on a single fortunate choice of hyperparameters. \textbf{No-Go} would mean the four groups' Pareto frontiers coincide within their confidence intervals, or that Group D underperforms Group C at matched error rate by an unacceptable margin (i.e.~that synchronicity itself, not verifier quality, is what erodes the benefit).

\textbf{Result}: on LmArena, the cross-encoder achieves a statistically significant hit-rate improvement at its best grid point (+1.9pp, non-overlapping confidence intervals), and under honest calibration (threshold chosen on the first half of the gray-zone stream, measured on the second, compared on the same requests) it beats the static threshold at all 6 tested τ\_low values (+0.87 to +3.10pp, Section 5.4). The result therefore does not rest on one lucky hyperparameter choice, and it clears the Go bar.

\textbf{But} this is a \textbf{weak Go}, for three reasons:

\begin{enumerate}
\tightlist
\item
  The lead is less than one-tenth of the oracle ceiling --- the real verifier is far from cashing in the mechanism's potential.
\item
  On the fixed threshold grid, 11 of 30 points were worse than the static threshold: the verifier threshold has to be calibrated per deployment (as honest calibration does) rather than taken from a default, which is itself an operational cost.
\item
  On SearchQueries, the same verifier is \textbf{net harmful}: 23 of 36 grid points lose to the static-threshold frontier, and under honest calibration, compared on the same requests, it loses at 5 of 6 points and ties at 1. Inside the default gray zone, raw similarity ranks candidates better than this untuned verifier does. Short keyword queries are far from the ms-marco training distribution (long web-search queries and passages), which is a plausible contributing cause but has not been isolated.
\end{enumerate}

So this paper's verdict is neither ``synchronous verification gating is worth doing'' nor ``it is not''. \textbf{The mechanism has real and often large headroom (the oracle results show it), but whether that headroom becomes a net production benefit depends on how well the verifier matches the data domain.} Prior work, Krites in particular, whose evaluation never used a real verifier, has not provided this conclusion. Section 5.6 shows that matching the domain is something a practitioner can do, not just a diagnosis: fine-tuning the same verifier on in-domain gray-zone labels turns SearchQueries, the dataset behind reason 3 above, from net harmful into a verifier that beats the static threshold at nearly every tested point.

\textbf{External cross-validation.} Baral et al.~(2026), an independent study from Redis and NYU, measure the same kind of gap: how much of a model's offline ranking quality survives as precision at a deployed operating point. Their retention ratio (CRR in v1, ORR in v4) divides the area under the deployed precision--cache-hit-ratio curve by offline PR-AUC. For \texttt{ms-marco-MiniLM-L12-v2}, a larger checkpoint from the same family as this paper's L6 model, they report 0.427 in v1 and 0.612 in v4, where it sits mid-pack. Their datasets (general paraphrase corpora), metric, and model all differ from this paper's, so the agreement is directional only: an off-the-shelf ms-marco cross-encoder can look reasonable offline and still lose much of its reliability at a fixed operating point. Baral et al.~stop at diagnosing the gap and calibrate the retrieval score itself; this paper adds a separate verification stage and tests a remedy (Section 5.6).

Applying their v1 formula to this paper's own scores gives retention ratios of 0.92--0.98 (SearchQueries Group D 0.979, Group E 0.924; LmArena Group D 0.982; Quora Group D 0.981), even though PR-AUC varies from 0.480 (Quora) to 0.938 (LmArena). This does not mean this paper's verifiers are better calibrated. The ratio is designed for retrieve-then-rerank pipelines, where the right candidate can score well offline but fail to reach first place after reranking; in this paper's single-tier design, similarity search commits to one candidate before the verifier runs, so that failure cannot occur and the ratio sits near its ceiling. Absolute values are not comparable across the two architectures.

\subsection{Relation to Krites' claim that synchronicity erodes benefit}\label{relation-to-krites-claim-that-synchronicity-erodes-benefit}

Krites predicts that synchronous verification ``will increase latency and erode the cache's benefit,'' but does not implement or measure it. Group C (oracle, synchronous) shows that, given an accurate verifier, synchronicity does not erode the hit-rate benefit: it delivers a large hit-rate improvement at matched error rate. What decides whether the net effect is positive is verifier quality (Groups D and E), a variable Krites' oracle-only evaluation cannot see. The latency cost is real, and its size depends on the verifier (Section 5.5): a co-located cross-encoder adds about 9--19ms per request at the highest call rates, while an LLM judge's cost depends on the model, provider and network path and was not measured here. Whether the cost is acceptable depends on a deployment's latency budget, which this paper cannot settle in general.

\section{Limitations}\label{limitations}

\begin{itemize}
\tightlist
\item
  \textbf{Correctness labels are noisy.} They come from the benchmarks' \texttt{equivalence\_\allowbreak{}id}, not from a direct judgment of whether an answer can be reused. Blinded human review (Section 5.15) estimates that 81--88\% of direct hits labelled wrong are actually reusable. Gray-zone noise varies by dataset: LmArena's is largely reliable, about 40\% of SearchQueries' gray-zone requests labelled wrong are reusable, and Quora cannot be checked because it has no real answers. Absolute error rates are therefore suitable for relative comparison only; comparisons at matched error rate were recomputed with denoised labels and are unaffected (Appendix A.3). One annotator did all the labelling, and no inter-annotator agreement was measured.
\item
  \textbf{Quora's equivalence classes are a reconstruction.} This paper built them by union-find over pairwise duplicate labels, and its \texttt{answer} field is a matched question's own text rather than a generated response (Section 5.6), so Quora is not constructed the same way as the other two datasets.
\item
  \textbf{All datasets are single-turn.} Each request is judged against a cached answer with no conversation history. Atalar et al.~(2026) report that embedding similarity predicts acceptable reuse poorly in multi-turn conversations (AUC 0.62--0.74), so these findings should not be assumed to carry over to multi-turn caching.
\item
  \textbf{Only the cross-encoder family was tested.} Section 5.12 swaps in a larger model and a more broadly trained reranker, and both stay within 0.03 AUC of the L6 model. NLI models, small generative LLMs, and hosted API judges remain untested as verifiers.
\item
  \textbf{Most operating points are chosen after the fact.} The grid-search results in Sections 5.3--5.6 report the best point of a threshold grid chosen per dataset and model. Honest calibration (Section 5.4) checks the main conclusions at 6 points per dataset, but not every table. Group A's frontier is also chosen after the fact, which makes comparisons conservative for the verifier; in the grid-search comparisons it is linearly interpolated between its nine grid points, which could over- or understate margins where the true frontier curves between them. The bootstrap resamples requests independently and ignores temporal correlation within the stream.
\item
  \textbf{The external cross-validation (Section 6.1) is directional only.} Baral et al.~evaluate a different checkpoint of the same family (\texttt{ms-marco-MiniLM-L12-v2}), on different datasets, with a different metric, and revised their numbers between versions.
\item
  \textbf{Latency was measured on few paths.} The cross-encoder was measured server-side and from one US-East client. Other regions, \texttt{/\allowbreak{}v1/\allowbreak{}verify/\allowbreak{}batch}, and latency under concurrent load are untested. The latency of a real LLM judge was not measured; the oracle's 70ms is a modeling assumption.
\item
  \textbf{Section 5.7's noise and drift are simulated.} Label noise is uniform random flips rather than a real feedback channel, whose errors may be systematic, and drift is simulated by slicing a static dataset into 8 chunks. A deployment would need to check the noise threshold and re-tuning cadence against its own traffic.
\item
  \textbf{Section 5.8's equivalence classes are an embedding-clustering proxy} with an untuned 0.92 cosine threshold, so its absolute numbers indicate the phenomenon rather than measure it precisely. The data is from 2014--2017, before chatbot-mediated support was common, and its CC0 source asks to be contacted for commercial use.
\item
  \textbf{Section 5.9's monitor was validated retrospectively} on a static held-out split of two datasets (one with the failure mode, one without), with detection thresholds chosen once and not tuned further.
\item
  \textbf{Section 5.10's verb extractor was not validated against human annotation}, so the share of its negative result due to extractor errors cannot be fully separated from the bucketing idea itself, although LmArena's regression has an extractor-independent cause. Only verb-alone keys were tested, not (verb, object) pairs.
\item
  \textbf{The Top-K cascade was tested only at K=2} and only with cross-encoder verifiers (Section 5.14). The latency of K sequential verifier calls was not measured, and on LmArena the ``K and fine-tuning are complementary'' comparison is less reliable than on the two datasets with finer threshold grids.
\item
  \textbf{This work does not address the problem Krites solves} (promoting curated static-tier answers into a dynamic tier); the two are complementary.
\end{itemize}

\section{Conclusion and Future Work}\label{conclusion-and-future-work}

In the single-tier, single-turn semantic caching settings evaluated here, putting a verifier synchronously on the serving path can raise hit rate by 20--28 percentage points at matched error rate when the verifier is perfect. An off-the-shelf cross-encoder, uncalibrated for the target domain, captures only a small slice of that headroom on LmArena and is net harmful on SearchQueries, under both post-hoc grid search and honest calibration on the same requests. This is the paper's core Go/No-Go finding (Section 6.1): a weak Go. The gap is not fundamental to the mechanism. Fine-tuning the same verifier on in-domain gray-zone labels, which a deployment would have to collect through audits or user feedback, makes it beat the static threshold at every honestly calibrated point on all three datasets (Sections 5.4 and 5.6). The recipe is reasonably robust (Section 5.7): it tolerates about 30\% label noise, needs up to about a thousand in-domain examples on the hardest dataset, and, after an initial drop, holds steady as traffic moves away from its training window. It is not unconditionally robust: on multi-year customer-support traffic from one brand, a drifting gray-zone base rate makes fine-tuning harmful (Section 5.8), and a simple change-point test on the labels detects that drift while staying silent on a stable brand (Section 5.9). The next question is therefore not whether to verify synchronously, but how to keep a verifier calibrated to a specific and possibly drifting domain. Each step of the path tested here, from fine-tuning to drift detection, still needs validation on more datasets and on live deployments.

Future work. Each item below is an open question left by a specific section. Questions the paper has already answered are reported in those sections and are not repeated here.

\textbf{Validation on real deployments}

\begin{enumerate}
\tightlist
\item
  Validate all conclusions, which currently come from trace-driven simulation, against real production traffic.
\item
  Validate the noise-tolerance and minimum-data figures of Section 5.7 against a real feedback channel (for example user thumbs-up/down rather than synthetic uniform label flips) and continuous production traffic, rather than resliced static datasets or, as in Section 5.8, a historical corpus with a known end date.
\item
  Run the drift monitor of Section 5.9 online, against a live and continuously rolling deployment. It has been validated only retrospectively, on two datasets.
\end{enumerate}

\textbf{The verifier}

\begin{enumerate}
\setcounter{enumi}{3}
\tightlist
\item
  Explain the remaining harm of the untuned verifier on SearchQueries: how much comes from short queries themselves and how much from factors not yet identified (Section 5.4 and Appendix A.1).
\item
  Test other verifier architectures, such as NLI models, small generative LLMs, and hosted LLM judges, including their real latency on the serving path (Sections 5.5 and 5.12).
\end{enumerate}

\textbf{Decision rule and candidate retrieval}

\begin{enumerate}
\setcounter{enumi}{5}
\tightlist
\item
  Top-K cascade (Section 5.14): whether K = 3--5 is worth the extra calls, and the cascade's real end-to-end latency under synchronous deployment.
\item
  Bucketing pre-filter (Section 5.10): whether keying on an (action, object) pair closes the blind spot that verb-only keys leave.
\item
  Rewrite instead of reject (Section 5.13): whether passing the rewritten answer through a second verification step recovers the negative result.
\end{enumerate}

\textbf{Scope}

\begin{enumerate}
\setcounter{enumi}{8}
\tightlist
\item
  Multi-turn conversations. All datasets here are single-turn (Section 7); Atalar et al.~(2026) show that reuse there depends on conversation history.
\end{enumerate}

\section*{Revision History}

Results are reported in their final form in the sections above; this list records when each change was made.

\begin{itemize}
\tightlist
\item
  \textbf{2026-07-26} First release.
\item
  \textbf{2026-08-09} SearchQueries data defect fixed and the affected results rerun (Appendix A.1).
\item
  \textbf{2026-08-15 to 08-17} Honest-calibration ablation (Section 5.4); Group B's port fixed to include vCache's two seed observations per cache entry, which raised its hit rate 4.4x--29.1x with the error rate still below δ (Section 5.2 reports only the fixed numbers); τ\_high sweep (Section 5.11).
\item
  \textbf{2026-08-22} Top-K cascade (Section 5.14).
\item
  \textbf{2026-09-03 to 09-09} Latency of the hosted verifier measured on the server, over the network, and from a US client (Section 5.5).
\item
  \textbf{2026-09-24} Group D/E compared with Group A on the same requests (Appendix A.2).
\item
  \textbf{2026-09-27} Label-noise correction; Section 5.11's core conclusion withdrawn (Appendix A.3).
\item
  \textbf{2026-11-25} Related work updated (Baral et al.~v4, Gill et al.~2025, Atalar et al.~2026); future work rewritten as open questions only; inline revision notes consolidated into this list; competing-interests statement added.
\item
  \textbf{2026-11-29 to 11-30} Related work extended (Liu et al.~2026; Atalar et al.~2026). Section 5.12 now compares all three models on the same gray-zone pairs; the earlier apparent LmArena gain of +0.13--0.15 AUC for the larger rerankers came from a sample-set mismatch.
\end{itemize}

\section*{Competing Interests}

Chengyou Xin develops and operates CacheVerifier (\url{https://www.cacheverifier.com}), a commercial hosted verification service built on the methods studied in this paper. Some latency measurements in Section 5.5 were made against that service. Where the text refers to ``the first author's commercial implementation'', it means this service; its code is not part of this paper's release. All benchmark experiments use the code and data in the accompanying public repository. The other author declares no competing interests.

\section*{Acknowledgments}

Groups A/B of this paper build directly on vCache's public benchmarks and reference implementation: the SemCacheLmArena and SemCacheSearchQueries datasets, the \texttt{STATIC\_\allowbreak{}THRESHOLDS}/\texttt{DELTAS} grids used in Groups A/B, and the \texttt{VerifiedDecisionPolicy} this paper ports line-by-line in Group B are all from the vCache project. This paper's evaluation framework would not have been possible without this public infrastructure. Thanks to the vCache authors --- L. G. Schroeder, A. Desai, A. Cuadron, K. Chu, S. Liu, M. Zhao, S. Krusche, A. Kemper, I. Stoica, M. Zaharia, and J. E. Gonzalez --- for open-sourcing it.

Sections 5.6 and 5.8's experiments likewise build directly on two other public datasets: Quora Question Pairs (Iyer, Dandekar, \& Csernai, 2017) and the Kaggle ``Customer Support on Twitter'' dataset (Axelbrooke, 2017). Thanks to the creators of both.

\section*{References}

\begin{itemize}
\tightlist
\item
  Bang, F. (2023). GPTCache: An open-source semantic cache for LLM applications enabling faster answers and cost savings. \emph{NLP-OSS 2023}.
\item
  Schroeder, L. G., Desai, A., Cuadron, A., Chu, K., Liu, S., Zhao, M., Krusche, S., Kemper, A., Stoica, I., Zaharia, M., \& Gonzalez, J. E. (2025). vCache: Verified semantic prompt caching. \emph{arXiv:2502.03771}.
\item
  Singh, A. K., Wang, H., Attaluri, S., Chiam, T., \& Zhu, W. (2026). Asynchronous Verified Semantic Caching for Tiered LLM Architectures. \emph{arXiv:2602.13165}.
\item
  Baral, A., Ralev, R., Zhechev, I. S., Rajamohan, S., \& Agarwal, J. (2026). Closing the Operational Gap in Semantic Caching. \emph{EMNLP 2026 Industry Track}; \emph{arXiv:2606.19719} (v4; v1, 18 Jun 2026, was titled ``Closing the Calibration Gap in Semantic Caching'').
\item
  Gill, W., Cechmanek, J., Hutcherson, T., Rajamohan, S., Agarwal, J., Gulzar, M. A., Singh, M., \& Dion, B. (2025). Advancing Semantic Caching for LLMs with Domain-Specific Embeddings and Synthetic Data. \emph{arXiv:2504.02268}.
\item
  Atalar, B., Beyene, S., Liu, X., \& Joe-Wong, C. (2026). Multi-Turn Semantic Caching for Low-Cost LLM Serving. \emph{arXiv:2611.01077}.
\item
  Liu, X., Atalar, B., Dai, X., Zuo, J., Wang, S., Lui, J. C. S., Chen, W., \& Joe-Wong, C. (2026). Semantic Caching for Low-Cost LLM Serving: From Offline Learning to Online Adaptation. \emph{IEEE INFOCOM 2026}; \emph{arXiv:2508.07675}.
\item
  Atalar, B., Liu, X., Zuo, J., Wang, S., Chen, W., \& Joe-Wong, C. (2026). Continuous Semantic Caching for Low-Cost LLM Serving. \emph{ACM MobiHoc 2026}; \emph{arXiv:2604.20021}.
\item
  Cheema, M. T., Aamir, A., Muhammad, K. G., Bhatti, N. A., Qazi, I. A., \& Qazi, Z. A. (2025). TweakLLM: A Routing Architecture for Dynamic Tailoring of Cached Responses. \emph{arXiv:2507.23674}.
\item
  Iyer, S., Dandekar, N., \& Csernai, K. (2017). First Quora dataset release: Question pairs. \emph{Quora Engineering Blog}.
\item
  Axelbrooke, S. (2017). Customer Support on Twitter. \emph{Kaggle}.
\item
  Xin, C. (2026a). Synchronous Online Verification Gating in Semantic Caches: An Empirical Study (technical report). \emph{Zenodo}. \url{https://doi.org/10.5281/zenodo.21703364}
\item
  Xin, C. (2026b). Finite-Sample Risk Control for Semantic Cache Reuse Decisions: Formal Guarantees, a Self-Selection Feedback Loop, and Economics. \emph{Zenodo}. \url{https://doi.org/10.5281/zenodo.22663724}
\item
  Xin, C. (2026c). Adversarial Robustness of Semantic Cache Verifiers: Gaps Exposed by Red-Teaming and Partial Repair via Training. \emph{Zenodo}. \url{https://doi.org/10.5281/zenodo.22661311}
\item
  Xin, C. (2026d). Stress-Testing Semantic Cache Reuse Decisions: Uncertainty Signals, a Discarded Similarity Signal, and Online Adaptive Thresholds. \emph{Zenodo}. \url{https://doi.org/10.5281/zenodo.22665784}
\end{itemize}

\appendix
\section{Corrections made after the original technical report}\label{appendix-a-corrections-made-after-the-original-technical-report}

The results in the main text are reported in their corrected form. This appendix records the three corrections that changed published numbers, what changed, and what did not.

\subsection{SearchQueries answer field was an unpopulated placeholder (2026-08-09)}\label{a.1-searchqueries-answer-field-was-an-unpopulated-placeholder-2026-08-09}

The SearchQueries dataset's Group D/E results had a data defect: the upstream vCache HuggingFace dataset (\texttt{vCache/\allowbreak{}SemBenchmarkSearchQueries})`s \texttt{response\_\allowbreak{}llama\_\allowbreak{}3\_\allowbreak{}8b} column was never actually populated for this split --- all 150,000 records' \texttt{answer} field was the identical constant placeholder string (\texttt{"Not\ required\ for\ the\ benchmark\ because\ of\ the\ id\_\allowbreak{}set"}), not the ``Llama-3-8B-generated answers'' this paper originally described. This meant the Group D/E cross-encoder verifier's second scoring input (the candidate answer text) was identical across every candidate and carried no discriminative information --- sufficient on its own to explain the previously reported ``near-chance'' verifier performance on SearchQueries, with no need for a more complex semantic explanation.

Real answers have been regenerated with the DeepSeek API (\texttt{deepseek-chat}; not Llama-3-8B, since no API access to that specific model was available) for the 103,188 affected records (the candidates actually read by the verifier in the gray zone, 68.8\% of the dataset), and the full Group D/E results --- including all downstream ablations (Sections 5.7/5.8) --- were rerun using the identical match trace and a threshold grid recalibrated to the corrected model's own score distribution (rather than reusing the old grid --- exactly the lesson this paper's own Section 5.4 already emphasized), on the same machine that produced the original Group E results. Every SearchQueries figure in the main text has been replaced with the corrected version; the original (defective) result files remain archived in \texttt{results/\allowbreak{}search\_\allowbreak{}queries\_\allowbreak{}groupD\_\allowbreak{}cross\_\allowbreak{}encoder.\allowbreak{}json} / \texttt{search\_\allowbreak{}queries\_\allowbreak{}groupE\_\allowbreak{}finetuned.\allowbreak{}json} for diffing. \textbf{The correction is not uniformly ``better.''} The fine-tuned Group E result is now stronger than originally reported, but the off-the-shelf Group D result is now worse --- the original data made it look merely ``close to ineffective''; the corrected data reveals it is genuinely worse than doing nothing at most operating points (details in Sections 5.4/5.6). Two of the three downstream ablations (label-noise robustness, cold-start data requirements) moved in the same direction: the placeholder data had systematically \emph{understated} these deployment risks. The third, drift, keeps its conclusion: the gain is lost mostly in the first step and then levels off. This defect does not affect any LmArena, Quora, or Twitter customer-support results --- none of them ever depended on the missing column.

\subsection{Honest-calibration comparisons were not made on the same requests (2026-09-24)}\label{a.2-honest-calibration-comparisons-were-not-made-on-the-same-requests-2026-09-24}

In the honest-calibration comparison at the end of §5.4 (and in the abstract, §6.1, §7 and §8, which reuse its numbers), the two sides were not computed on the same set of requests: the hit/error rates of each Group D/E honest operating point count only ``every request outside the gray zone + the second (test) half of the gray-zone stream'', while Group A's static-threshold frontier was built on all requests. The gray zone is where errors concentrate, so the comparison mixed in an ``easier or harder evaluation set'' effect, whose direction differs by dataset. The re-check first reproduces the published numbers exactly (\texttt{scripts/\allowbreak{}groupD\_\allowbreak{}vs\_\allowbreak{}groupA\_\allowbreak{}fair\_\allowbreak{}frontier.\allowbreak{}py}), then recomputes Group A's frontier on \textbf{exactly the same request set} as each D/E point (every distinct similarity value as a threshold; win/tie/loss by whether the bootstrap 95\% CI of the difference crosses 0, 300 request-level resamples; \texttt{scripts/\allowbreak{}groupD\_\allowbreak{}vs\_\allowbreak{}groupA\_\allowbreak{}fair\_\allowbreak{}frontier\_\allowbreak{}bootstrap.\allowbreak{}py}, results in \texttt{results/\allowbreak{}groupD\_\allowbreak{}vs\_\allowbreak{}groupA\_\allowbreak{}fair\_\allowbreak{}frontier*.\allowbreak{}json}). After the fix: (1) \textbf{the untuned verifier's honest-calibration result on SearchQueries changes from ``6/6 beat'' to 0 wins, 1 tie, 5 losses (−0.01 to −1.72pp)} --- the ``net harmful'' verdict §5.4 first obtained from post-hoc grid search still holds under honest calibration, and the earlier claim that it ``reverses'' is withdrawn; (2) the fine-tuned verifier (Group E) significantly beats the static threshold at all 18 tested points across the three datasets, with changed magnitudes: SearchQueries' best lead falls from +7.74pp to +2.33pp, LmArena's rises from +5.66pp to +6.92pp, Quora's rises from +0.43pp to +0.87pp; (3) the cost-ratio analysis in the companion paper on formal risk control (Xin, 2026b), which reused the same numbers, has been corrected in the same way. The corresponding numbers and conclusions in the main text have been replaced; the original numbers are kept in \texttt{results/\allowbreak{}*honest\_\allowbreak{}calibration*.\allowbreak{}json} and \texttt{results/\allowbreak{}cost\_\allowbreak{}sensitive\_\allowbreak{}reanalysis.\allowbreak{}json}. The paper's core verdict (weak Go; in-domain fine-tuning is the main lever that turns the mechanism into a net gain) is unaffected and in fact more consistent: the untuned verifier beats the static threshold only on LmArena, where it is itself strong. The post-hoc grid-search comparisons (first half of §5.4, §5.6) compute both sides on all requests and are not affected; re-checking them with a dense frontier instead of 9-point linear interpolation leaves their qualitative conclusions unchanged (\texttt{results/\allowbreak{}grid\_\allowbreak{}search\_\allowbreak{}vs\_\allowbreak{}dense\_\allowbreak{}groupA\_\allowbreak{}check.\allowbreak{}json}). The same re-check also found that inside the default gray zone, ranking by raw similarity beats the untuned cross-encoder at every error budget on SearchQueries and Quora (\texttt{scripts/\allowbreak{}gray\_\allowbreak{}zone\_\allowbreak{}high\_\allowbreak{}precision\_\allowbreak{}compare.\allowbreak{}py}), consistent with the correction above.

\subsection{Benchmark correctness labels are noisy in the direct-hit zone (2026-09-27)}\label{a.3-benchmark-correctness-labels-are-noisy-in-the-direct-hit-zone-2026-09-27}

Every ``was this reuse correct'' judgment in this paper comes from the dataset's own \texttt{equivalence\_\allowbreak{}id}: a reuse counts as correct only if the new request and the reused entry belong to the same equivalence class. A blinded human review found that this label is heavily noisy in the direct-hit zone (similarity ≥ 0.97). Among sampled direct hits labelled wrong in the three datasets, the annotator judged that the cached answer could be served as-is for 88\% of them (66/75, 95\% CI 79--94\%). Typical cases are the same question with different wording, word order or plural form (e.g.~``citi double cash credit card login'' vs.~``citi double cash login credit card''), plus a few samples whose two English questions are character-for-character identical yet were labelled as different classes. Combined with the LLM pre-classification shares, an estimated 81--88\% of the direct hits labelled wrong are actually reusable. Gray-zone (0.80--0.97) noise differs a lot by dataset: in a random blinded sample of 150, the share ``labelled wrong but reusable'' is 12\% on LmArena (CI 4--30\%) with 4\% ``labelled correct but not reusable''; on SearchQueries the two shares are 40\% (CI 23--59\%) and 12\%; Quora has no real answers (its answer field is the question itself) and cannot be checked this way. One annotator did all labelling; no inter-annotator agreement was measured. Method and full results are in Section 5.15.

Re-scoring with denoised labels (an expected-value correction in the direct-hit zone at the estimated noise rate; gray-zone labels unchanged):

(1) \textbf{Unaffected}: every ``at matched error rate'' comparison --- the hit-rate gap of synchronous verification (Groups D/E) and of the oracle verifier (Group C) over the static threshold (Group A) moves by at most about 0.2 percentage points on all three datasets. Both sides serve exactly the same direct hits, and the matched Group A frontier points almost all lie at τ \textless{} 0.97, so the noise cancels. The weak-Go verdict, the effectiveness of in-domain fine-tuning, and the net harm of the untuned verifier on SearchQueries therefore stand. All CRC results in the companion paper on formal risk control (Xin, 2026b) count gray-zone risk only and are not affected by direct-hit noise; LmArena's gray-zone labels were verified to be largely reliable.

(2) \textbf{Absolute error rates are systematically overstated}, more so the larger the share of direct-hit errors at an operating point. Denoised point estimates: on LmArena, Groups D/E go from about 5.6--6.4\% to about 1.1--1.5\%, and Group A at τ = 0.97 from 5.2\% to 0.95\%; on SearchQueries, Group E goes from 5.5--13\% to 1.1--7.2\%; on Quora, Group A at τ = 0.97 goes from 1.05\% to 0.20\%. Absolute error rates in the text should be read as benchmark-label figures, suitable for relative comparison only.

(3) \textbf{The core conclusion of Section 5.11 is withdrawn.} The claim ``on LmArena, raising τ\_high from 0.97 to 0.99 raises the best net lead from +1.42pp to +5.18pp (24/30 points win)'' is an artifact of label noise. At τ\_high = 0.99 the verifier judges candidates in {[}0.97, 0.99); rejecting the ones labelled wrong looks like avoiding errors under the noisy labels, but about 80\% of them are actually reusable, so rejecting them only loses hits. After denoising, LmArena at τ\_high = 0.99 has 0/30 wins with a best lead of −1.24pp (+0.03 / −1.86pp at the two ends of the noise-rate interval), and its best τ\_high is 0.97 (15/30 wins, best +1.87pp); SearchQueries at τ\_high = 0.99 worsens from −0.11pp to −1.46pp; Quora is unchanged. After denoising, none of the three datasets supports raising τ\_high to 0.99. (The {[}0.97, 0.99) band was re-scored with the original model and all 285 operating points of Section 5.11 were first reproduced exactly; this also showed that the ``5/30 win'' previously reported for SearchQueries at τ\_high = 0.95 should read 2/30.)

(4) \textbf{The mid-to-high-r end of the cost-sensitive analysis (Xin, 2026b, Section 4.3) changes}: Group A's conservative thresholds take all their errors from the direct-hit zone, become much cheaper after denoising, and open a middle range where the static threshold wins; see that paper's 2026-09-27 erratum.

(5) \textbf{The SearchQueries interpretation needs a qualifier.} About 40\% of the SearchQueries gray-zone requests labelled wrong are actually reusable, so the verifier's low AUC on this dataset (0.5984 untuned, 0.7120 fine-tuned) is partly depressed by label noise rather than being purely a verifier weakness; CRC thresholds on this dataset are correspondingly conservative. Quora's gray-zone conclusions cannot be checked this way because the dataset has no answers.

(6) \textbf{Section 5.2 (vCache)}: per-request results were not kept, so this cannot be recomputed exactly. The direction is clear: its error rates can only be lower than reported, and since vCache learns its per-entry thresholds from these noisy labels, it would approve more under clean labels, so the reported hit rates are lower bounds. The conclusion ``error rate stays below the target δ'' is unaffected.

Scripts and results: \texttt{scripts/\allowbreak{}direct\_\allowbreak{}hit\_\allowbreak{}error\_\allowbreak{}taxonomy.\allowbreak{}py}, \texttt{scripts/\allowbreak{}denoised\_\allowbreak{}rescore.\allowbreak{}py}, \texttt{scripts/\allowbreak{}denoised\_\allowbreak{}cost\_\allowbreak{}sensitive.\allowbreak{}py}, \texttt{scripts/\allowbreak{}denoised\_\allowbreak{}tau\_\allowbreak{}high.\allowbreak{}py}, \texttt{scripts/\allowbreak{}gray\_\allowbreak{}zone\_\allowbreak{}label\_\allowbreak{}check.\allowbreak{}py}; \texttt{results/\allowbreak{}direct\_\allowbreak{}hit\_\allowbreak{}annotation\_\allowbreak{}*}, \texttt{results/\allowbreak{}gray\_\allowbreak{}zone\_\allowbreak{}annotation\_\allowbreak{}*}, \texttt{results/\allowbreak{}denoised\_\allowbreak{}*.\allowbreak{}json}.

\end{document}